\documentclass[10pt]{article} %
\usepackage[preprint]{tmlr}

\usepackage{amsmath,amsfonts,bm}

\def\eqref#1{equation~\ref{#1}}

\def\1{\bm{1}}

\def\vb{{\bm{b}}}
\def\vc{{\bm{c}}}

\def\vp{{\bm{p}}}

\def\vz{{\bm{z}}}

\def\mA{{\bm{A}}}

\def\mI{{\bm{I}}}

\def\mM{{\bm{M}}}

\def\mV{{\bm{V}}}

\DeclareMathAlphabet{\mathsfit}{\encodingdefault}{\sfdefault}{m}{sl}
\SetMathAlphabet{\mathsfit}{bold}{\encodingdefault}{\sfdefault}{bx}{n}

\usepackage{soul}

\soulregister\mM{0}
\soulregister\mathcal{0}
\soulregister\cref{0}
\soulregister\textbf{0}

\definecolor{turquoise}{cmyk}{0.65,0,0.1,0.3}
\definecolor{purple}{rgb}{0.65,0,0.65}
\definecolor{dark_purple}{rgb}{0.45,0,0.45}
\definecolor{dark_green}{rgb}{0, 0.5, 0}
\definecolor{orange}{rgb}{0.8, 0.6, 0.2}
\definecolor{red}{rgb}{0.8, 0.2, 0.2}
\definecolor{darkred}{rgb}{0.6, 0.1, 0.05}
\definecolor{light_gray}{rgb}{0.7, 0.7, .7}
\definecolor{pink}{rgb}{0.9, 0, 0.6}
\definecolor{greyblue}{rgb}{0.25, 0.25, 1}
\definecolor{teal}{rgb}{0.0, 0.4, 0.4}
\definecolor{texthighlight}{rgb}{1.0,0.85,0.3} %

\newcommand{\da}[1]{{\color{magenta}#1}}

\def \customparskip {0.2em}
\renewcommand{\paragraph}[1]{\vspace{\customparskip}\noindent\textbf{#1}}

\sethlcolor{texthighlight} %

\renewcommand{\da}[1]{#1}

\usepackage{makecell}
\usepackage{lipsum}
\usepackage{xspace}
\usepackage{caption} 
\usepackage{subcaption} 
\usepackage{booktabs}
\usepackage{enumitem}

\usepackage{amssymb} %
\usepackage{algorithm} %
\usepackage{algpseudocode} %
\usepackage{pifont} %
\usepackage[normalem]{ulem} %

\newcommand{\sig}{\operatorname{Sig}} %

\newcommand{\trailblazer}{Trailblazer~\cite{ma2024trailblazer}\xspace}
\newcommand{\turbo}{T2V-Turbo~\cite{li2024t2v}\xspace}
\newcommand{\peekaboo}{Peekaboo~\cite{jain2024peekaboo}\xspace}
\newcommand{\freetraj}{Freetraj~\cite{qiu2024freetraj}\xspace}

\newcommand{\hparam}[1]{{\lambda_{\text{#1}}}}
\newcommand{\loss}[1]{{\mathcal{L}_{\text{#1}}}}
\newcommand{\ie}{i.e.\xspace}
\newcommand{\etal}{\textit{et.al.}\xspace}
\newcommand{\eg}{e.g.\xspace}

\usepackage[table]{xcolor}
\definecolor{bestcell}{RGB}{170, 235, 170} %
\definecolor{secondbestcell}{RGB}{230, 255, 230} %

\let\titleold\title
\renewcommand{\title}[1]{\titleold{#1}\newcommand{\thetitle}{#1}}
\def\maketitlesupplementary
   {
   \clearpage
   \begin{center}
   \vspace*{0.5em}
   \textbf{\LARGE Supplementary Material}
   \vspace{1.0em}
   \end{center}
   }

\usepackage{hyperref}
\usepackage{url}
\usepackage{graphicx}
\usepackage[capitalize,noabbrev,nameinlink]{cleveref}

\title{Making Video Models Adhere to User Intent with Minor Adjustments}

\author{\da{\name Daniel Ajisafe \email dajisafe@cs.ubc.ca \\
      \addr Department of Computer Science\\
      The University of British Columbia
      \AND
      \name Eric Hedlin \email iamerich@cs.ubc.ca \\
      \addr Department of Computer Science\\
      The University of British Columbia
      \AND
      \name Helge Rhodin \email helge.rhodin@uni-bielefeld.de \\
      \addr Faculty of Technology\\
      Bielefeld University \\
      \addr Department of Computer Science\\
      The University of British Columbia
      \AND
      \name Kwang Moo Yi \email kmyi@cs.ubc.ca\\
      \addr Department of Computer Science\\
      The University of British Columbia
      }
      }

\def\month{03}  %
\def\year{2026} %
\def\openreview{\url{https://openreview.net/forum?id=Opvq2wfBR5}} %

\begin{document}
{%
\renewcommand\twocolumn[1][]{#1}%
\maketitle
\newcommand{\vtextheight}{3cm}
\newcommand{\teaserfigw}{0.23}
\newcommand{\teaserfigh}{0.21}
\newcommand{\teasertextw}{0.23}
\newcommand{\colwidthA}{0.265\linewidth}
\newcommand{\colwidthB}{0.265\linewidth}
\newcommand{\colwidthC}{0.213\linewidth}
\newcommand{\colwidthD}{0.213\linewidth}
\setlength{\tabcolsep}{1pt} 
\begin{tabular}{c p{\colwidthA} p{\colwidthB} p{\colwidthC} p{\colwidthD}}
    {\rotatebox[origin=l    ]{90}{\parbox[b]{\vtextheight}{\centering Baseline}}} &
    \includegraphics[height=\teaserfigh\textwidth]{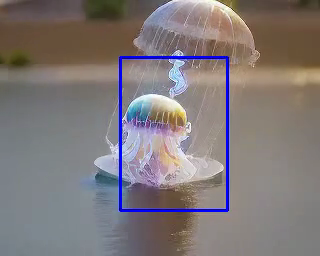} &  
    \includegraphics[height=\teaserfigh\textwidth]{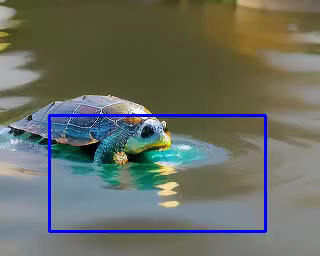} &  
    \includegraphics[height=\teaserfigh\textwidth]{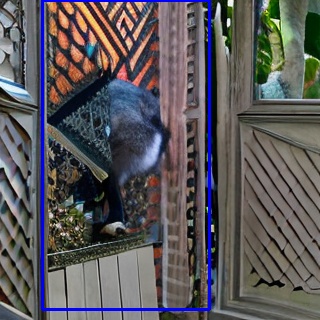} &  
    \includegraphics[height=\teaserfigh\textwidth]{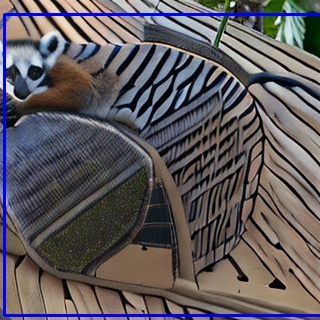}  
     \\
{\rotatebox[origin=l]{90}{\parbox[b]{\vtextheight}{\bf \centering Our method}}} &
    \includegraphics[height=\teaserfigh\textwidth]{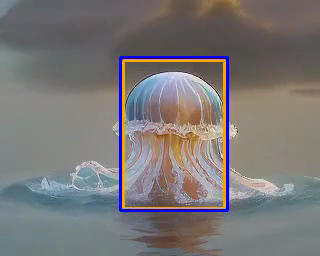} &  
    \includegraphics[height=\teaserfigh\textwidth]{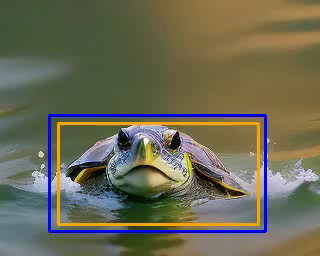} &  
    \includegraphics[height=\teaserfigh\textwidth]{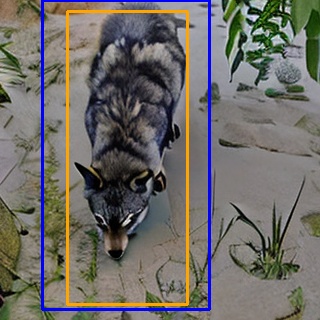} &  
    \includegraphics[height=\teaserfigh\textwidth]{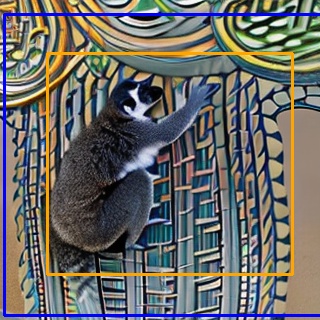} 
     \\
     &
     \parbox[t]{\linewidth}{\small  \centering \footnotesize\texttt{The jellyfish is swimming}} & 
     \parbox[t]{\linewidth}{\small \centering \footnotesize\texttt{The turtle is swimming}} & 
     \parbox[t]{\linewidth}{\small \centering \footnotesize\texttt{The wolf is exploring}} & 
     \parbox[t]{\linewidth}{\small \centering \footnotesize\texttt{The ring-tailed lemur is exploring}} 
\end{tabular}

\captionof{figure}{
    {\bf Teaser -- }
    We show example bounding box controlled video generations with (left) \trailblazer in \turbo and (right) the original \trailblazer respectively. On top is the 
    original control signal and below is
    our adjusted bounding boxes.
    We show the original bounding boxes as \textcolor{blue}{blue} and the adjusted boxes as \textcolor{orange}{orange}.
    While the modification is subtle, the difference in the quality of generation is large.
    We modify bounding boxes to adhere better to the cross-attention maps within the video models.
}
\vspace{1em}
\label{fig:teaser}
}

\begin{abstract}
With the recent drastic advancements in text-to-video diffusion models, controlling their generations has drawn interest.
A popular way for control is through bounding boxes or layouts.
However, enforcing adherence to these control inputs is still an open problem.
In this work, we show that by slightly adjusting user-provided bounding boxes we can improve both the quality of generations and the adherence to the control inputs.
This is achieved by simply optimizing the bounding boxes to better align with the internal attention maps of the video diffusion model while carefully balancing the focus on foreground and background.
In a sense, we are modifying the bounding boxes to be at places where the model is familiar with.
Surprisingly, we find that even with small modifications, the quality of generations can vary significantly.
To do so, we propose a smooth mask to make the bounding box position differentiable and an attention-maximization objective that we use to alter the bounding boxes.
We conduct thorough experiments, including a user study to validate the effectiveness of our method. Our code is made available on the 
\href{https://ubc-vision.github.io/MinorAdjustVideo/index.html}{project webpage}
to foster future research from the community.

\end{abstract}
    
\section{Introduction}
\label{sec:intro}

Text-to-video diffusion models have made groundbreaking advances in producing high-quality prompt-directed generations~\cite{ho2022video, singer2023make, weissenborn2020scaling, arnab2021vivit, 
ho2022imagen, chen2024videocrafter2}.
Among the various directions, methods based on diffusion~\cite{ho2022video, ho2022imagen, chen2024videocrafter2} and transformers~\cite{weissenborn2020scaling, arnab2021vivit} have become popular.
While these models can be controlled through proper prompting, this is not always straightforward and requires careful prompt engineering~\cite{liu2023pre}. In particular, the spatial control of object placement and object trajectories remains difficult.

Naturally, researchers have sought to improve the controllability of text-to-video diffusion models.
These include methods that specifically train a model in addition to the main model~\cite{controlnet}, which was shown to be especially effective for text-to-image generation~\cite{wang2023videocomposer, controlnet, patashnik2021styleclip}.
For video generation, however, training such additional models is computationally expensive.
For spatial control, 
using bounding boxes or layouts as control inputs~\cite{ma2024trailblazer, zheng2023layoutdiffusion} has gained attention.
These approaches work without additional training by simply modifying the internal attention maps within the video diffusion model~\cite{ma2024trailblazer, hertz2023prompt, chefer2023attend, tumanyan2023plug} or through guidance~\cite{patashnik2021styleclip, nichol2022glide} with a pre-trained classifier.
While these methods are effective, they are still limited in terms of adherence to the control inputs.
Generated videos contain artificial outcomes because of the mismatch between how the control signals affect the video generation process and how it was trained without such injection; see \cref{fig:teaser}.

In this work, we show that by slightly adjusting the control inputs, we can 
improve both the quality of generations and the control adherence. 
However, finding how to adjust bounding boxes to adhere to user control without hurting the final generation outcome is challenging. For instance, modifying and balancing attention within the diffusion model is effective for control but can lead to over-saturation. In general,
changes that are outside of the model's internal understanding easily cause video generation to degrade.
To address these, we optimize bounding boxes to better align with the internal attention maps of the video diffusion model, while still being close to the user-provided bounding boxes.
More specifically, we introduce an optimization framework that ensures that attention maps edits remain differentiable with respect to the bounding box parameters.
With the differentiable pipeline, for improved alignment of the bounding box control and the video model, we then propose to maximize the attention within the bounding box of the \emph{next layer after the edit}, which is representative of where the neural network is focusing on once the edits are applied. 
Further, while we enhance focus, we also balance attention between the foreground and background areas, so that generation process does not completely ignore the background.
While the resulting adjustments to the bounding boxes are small, their impact on the video generation is significant; see \cref{fig:teaser}.

To summarize, we make the following contributions:
\begin{itemize}[itemsep=1pt, topsep=2pt, parsep=0pt]
    \item we demonstrate that small adjustments to user intent can lead to significant improvements in controlled video generation when adjusting with our method;
    \item we propose a novel method to optimize the control inputs to align well with the internal attention maps of the video diffusion model, while still being close to the user-provided bounding boxes; 
    \item to do so, we present an editing pipeline that alters a non-differentiable method to be differentiable with respect to the bounding box parameters;
    \item to find better bounding-box parameters, we propose a balanced attention maximization objective considering the cross-attention maps of the next layer after the edits; and 
    \item we conduct experiments including a user study to validate the effectiveness of our method, and show that it outperforms existing methods in terms of video generation quality and adherence to user intent.
\end{itemize}

\section{Related Work}

\vspace{-\customparskip}
\paragraph{Text-to-image generative models.}
Generative text-to-image models have shown ground-breaking results in terms of high-quality images~\cite{rombach2022high, nichol2022glide} synthesizing different objects, persons, and places. However, these models have been shown to fail to adhere faithfully to spatial user intent~\cite{rombach2022high, ramesh2022hierarchical, nichol2022glide}, hence requiring a separate modality e.g., boxes, scribble, etc., for better text-to-image alignment.

\paragraph{Text-to-image generative models with box control.}
With alternate input control such as simple user-defined boxes, several methods~\cite{zheng2023layoutdiffusion, chang2023muse} have demonstrated that the spatial composition can be controlled more faithfully. For example, Directed diffusion~\cite{ma2024directed} directs the placements of objects by introducing activations at desired positions in a text-to-image diffusion model, leading to a better generation outcome. BoxDiff~\cite{xie2023boxdiff} introduces spatial constraints such as inner-box, outer-box, and corner constraints in an optimization framework, for controlling objects and context in generated images. These methods follow a forward guidance approach. Alternatively, other work
~\cite{chen2024training} demonstrates the superiority of backward guidance over forward guidance for robust layout control. Their method optimizes the latent, allowing both guided and unguided tokens to influence the generation outcome. However, these methods are limited to single images and do not investigate their effectiveness on temporal data, i.e., video.

\begin{figure*}
    \centering
    \includegraphics[width=\textwidth]{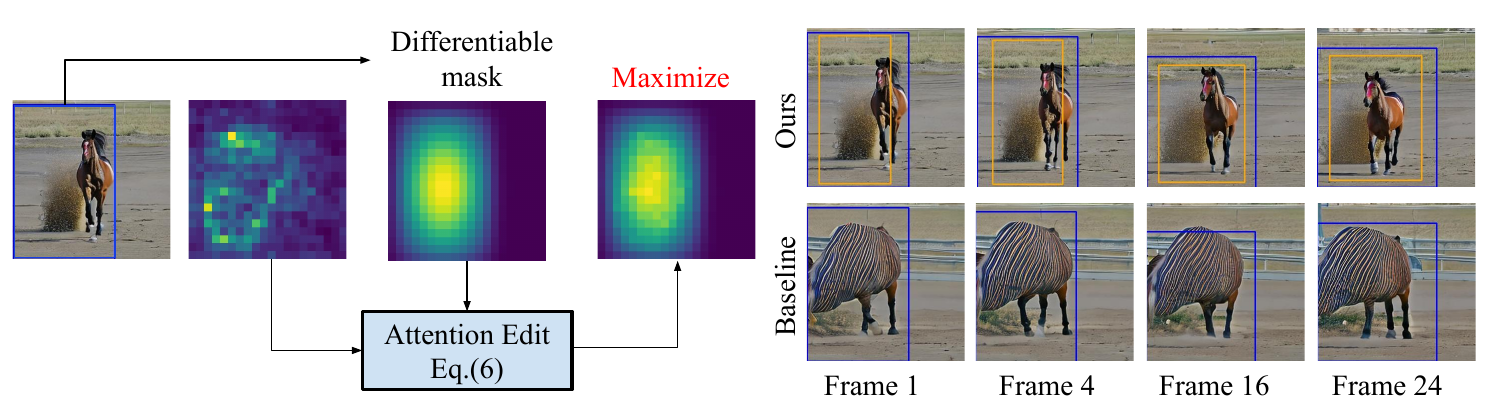}
    \caption{
        {\bf Overview -- }
        We inject bounding box control for video diffusion models by editing their cross attention maps within the network.
        However, not all such edits are friendly to video diffusion models as they are not trained with such edits.
        Thus, when applying these edits, we make sure that this editing process is differentiable (\cref{sec:edit}) and adjust the edit parameters in a way such that the network behaves as intended---attention being focused on desired regions 
        (\cref{sec:opt}).
        We show the original bounding boxes in \textcolor{blue}{blue} and the adjusted bounding boxes in \textcolor{orange}{orange}.  
        Though the adjustments are minimal and close to the original user input, they create a drastic difference in terms of video generation quality and adherence to the bounding boxes. 
    }
    \label{fig:overview}
\end{figure*}

\paragraph{Text-to-video generative models with box control.} 
There are several works that investigate the adherence to input control for videos~\cite{wang2023videocomposer, lian2024llm, jain2024peekaboo, ma2024trailblazer, wang2024boximator, chen2025motion, luo2025ctrl, qiu2024freetraj, lei2025ditraj}.
None, however, look into how slight changes in controls can lead to diverse final generation outcomes. 
\peekaboo uses attention masking to guide a video generation process, therefore utilizing local context for generating individual objects. 
However, their use of an infinite attention injection in the background often results in missing background details~\cite{ma2024trailblazer}.
\trailblazer, on the other hand, uses a direct/in-place replacement strategy to bias cross-attention maps towards user intent. With keyframed boxes, their method achieves controllable generation but often sacrifices better generation outcomes for more control. In other words, controlled generation outcomes are less faithful to the given prompt. 
Their method also relies on tuning hyper-parameters per object, which is not scalable. 
\da{\freetraj is also a training-free method that achieves controllable video generation by imposing guidance on both noise construction and attention computation. While it achieves good results, its trajectory injection sometimes leads to poor visual quality or poor trajectory alignment. 

Other methods, such as Motion-Zero~\cite{chen2025motion} and Boximator~\cite{wang2024boximator}, \da{do} not release implementation code or evaluation data, making fair experimental comparison difficult. Ctrl-V requires task-specific training and conditions video generation on an initial frame, whereas our method is training-free and operates directly on text-to-video diffusion backbones. Generally, other methods also differ substantially in supervision regime or underlying model architecture, which limits fair comparison.}

To the best of our knowledge, we are the first to explore the benefit of adjustments to box-controlled generation outcomes.  We build on the work of \trailblazer, and apply our method to two text-to-video models~\cite{ma2024trailblazer, li2024t2v}. 
We demonstrate our core contributions in the following sections.

\section{Method}
\label{sec:method}

Our core idea is to modify user input, \ie, the bounding boxes,
aligning them to the internal attention mechanisms within the video diffusion model.
As shown in \cref{fig:overview}, we implement this by introducing a differentiable attention map editing method, which we then optimize through to adjust the bounding boxes.
When adjusting, as we do not have any specific measure for what a good trajectory is---after all, we are generating a video from scratch---we look into how the attention maps evolve through the network layers. 
Specifically, we encourage the edits to be inline with what how the neural network creates further attention maps, that is, we regularize it such that the attention map, after editing, causes the attention map of the next layer within the neural network to focus within the box.

To explain our method, and for completeness, we first review how control injection is done for a training-free baseline, then discuss our work.

\subsection{Preliminary: \trailblazer}

Our goal is to generate a video with $F$ frames, where a desired object of interest, e.g., "cat", adheres faithfully to the motion and control of the bounding boxes $\mathcal{B} \in \mathbb{R}^{F \times 4}$ provided by the user. We build upon \trailblazer, which `injects' user control in the form of bounding boxes, by directly adjusting the internal attention maps of the video diffusion model.
A strong benefit in doing so is that there is no need to train or fine-tune the video diffusion model, and any off-the-shelf video diffusion model can be used as long as it has cross-attention layers.

Specifically,
at inference, we feed an input text embedding $\vp$, noisy latent code $\vz$, and timestep $t$, as input to a video diffusion model $\Theta_b$. 
After each cross-attention layer, we extract and edit cross-attention maps $\mA_{S} \in \mathbb{R}^{C \times H \times W \times N}$, temporal cross-attention maps $\mA_{T} \in \mathbb{R}^{C \times H \times W \times F \times F}$. 
Then, denoting the predicted noise sample as $\epsilon_{t}$, we can write
\begin{equation}
    \mA_{S}, \mA_{T}, \epsilon_{t} = \Theta_b(\vp, \vz, t)
    ,
\end{equation}
where $N$ is the number of text tokens, and $N=77$ when using CLIP tokenizer~\cite{radford2021learning}.  
Here, the control signal is then injected by modifying the attention maps $\mA_{S}$ and $\mA_{T}$.
Note here that $\mA_{S}$ encodes the relationship between the feature $\mV$ and text embedding $\vp$, while $\mA_{T}$ encodes the relationship between pixels among different frames, without any association to the text.

\begin{figure*}
    \centering
    \newcommand{\subfigsize}{0.245\linewidth}
    \begin{subfigure}[b]{\subfigsize}
        \includegraphics[width=\linewidth]{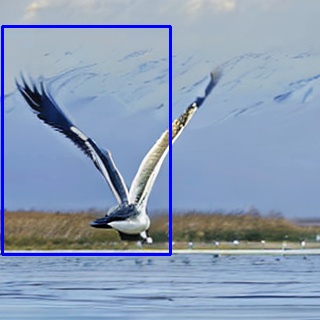}
        \caption{Generated frame and user control}
    \end{subfigure}
    \begin{subfigure}[b]{\subfigsize}
        \includegraphics[width=\linewidth]{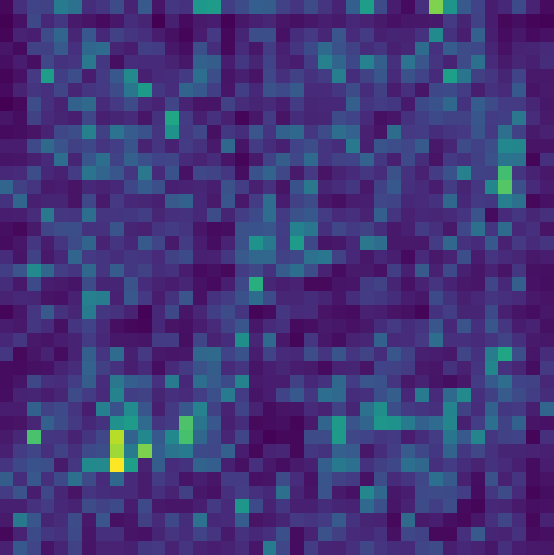}
        \caption{Attention map for the token `Swan'}
    \end{subfigure}
    \begin{subfigure}[b]{\subfigsize}
        \includegraphics[width=\linewidth]{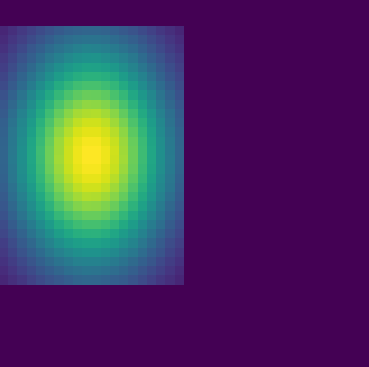}
        \caption{Attention map edited by \trailblazer}
    \end{subfigure}
    \begin{subfigure}[b]{\subfigsize}
        \includegraphics[width=\linewidth]{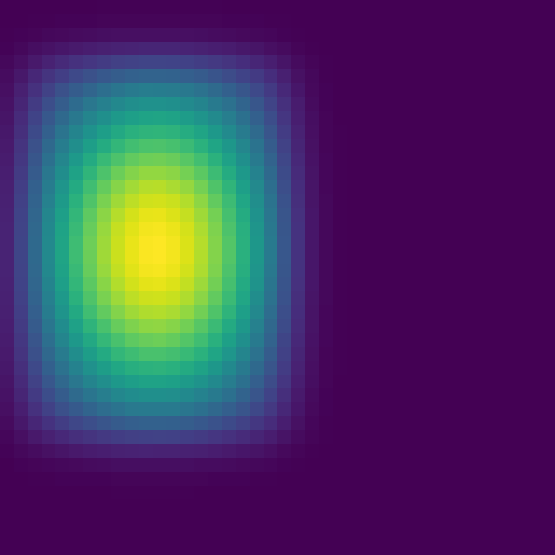}
        \caption{Attention map edited by our method}
    \end{subfigure}
    \caption{
        {\bf Example of attention map editing -- }
            We show an example of the generated video frame with the desired user control bounding box, and the associated attention map edits.
            While effective, \trailblazer{} relies on a replacement operation that is not differentiable with respect to the box parameters.
            Our method, on the other hand, performs a smooth differentiable edit.
        }
    \label{fig:mask_edit}
\end{figure*}

In more detail, the attention maps are modified by directly weakening the attention map outside of the bounding box via multiplying a constant weakening factor, and strengthening the attention map within the box by adding a Gaussian map.
Consider now the masks $\mM_{\mathcal{B}_{S}} \in \mathbb{R}^{C \times H \times W \times N} \in [0,1]$ and $\mM_{\mathcal{B}_{T}} \in \mathbb{R}^{C \times H \times W \times F \times F} \in [0,1]$ for the spatial and temporal layers, respectively, which encodes whether the pixels involved in the attention maps are either
within (1) the control bounding box or not (0).
These maps $\mM_{\mathcal{B}}$ are anchored by box parameters $\vb \in (b_l, b_t, b_r, b_b) \in \mathbb{R}^{4}$ 
representing top, left, right and bottom coordinates. $C, H, W$ stands for the number of channels, the height, and the width respectively. 
\trailblazer then modifies the attention maps as
\begin{equation}
    \begin{aligned}
        \bar{\mA}_{*} &= \underbrace{\mA_{*} \odot \left(\hparam{w}(1 - \mM_{*}) + \mM_{*}\right)}_{\text{weakening outside}} + \underbrace{\hparam{s}\mM_{\mathcal{G}} \odot \mM_{*}}_{\text{strengthening inside}}
    \end{aligned}
    ,
    \label{eq:trailblazer_attention_injection}
\end{equation}
where subscript $*$ denotes both spatial and temporal, $\odot$ is the element-wise multiplication, $\hparam{w}$ is the weakening factor which we set to 
$\hparam{w}{=}0.001$, and $\hparam{s}$ is the strengthening factor which we set to $\hparam{s}{=}0.15$,\footnote{\trailblazer uses a per-animal hyperparameter setting, but this is impractical for our evaluation scenario with hundreds of animals. We thus use this value to a fixed value that we found empirically and use it for all evaluations, including our method.} and $\mM_{\mathcal{G}}$ is a Gaussian map, defined as\footnote{In the case of the temporal attention map, there are two centers, one for each frame, and we simply take the maximum value for each pixel for overlapping regions.}
\begin{equation}
    \mM_{\mathcal{G}} = \exp\left(-\frac{\left(\mI_{coord}-{\vc}\right)^2}{2\boldsymbol{\sigma^2}}\right),
    \label{eq:gaussian_function}
\end{equation}
where $I_{coord}$ is a tensor of image coordinates, 
$\vc = \left(\frac{bl+br}{2}, \frac{bt+bb}{2} \right) \in \mathbb{R}^{2}$
is the center of the box, 
and standard deviation $\boldsymbol{\sigma} \in \mathbb{R}^{2}$ is typically set to one-third of the box height $h$ and width $w$.\footnote{The original paper claims one-half, but the official code release uses one-third, which is what we use.}

While effective as demonstrated in \citet{ma2024trailblazer}, the edit in \cref{eq:trailblazer_attention_injection} is not differentiable with respect to the box parameters as it contains discrete borders, especially where the mask $\mM_{*}$ transitions either from 0 to 1 or vice versa.
Moreover, the shape of the edit, as shown in \cref{fig:mask_edit}, has sharp discontinuities at the edges.
While the Gaussian map allows focusing the attention map to the center of the box, the standard deviation $\sigma$ of the Gaussian is set to a large enough value to ensure that the attention map is spread over the entire box, but this then leads to a truncated shape with clipped edges.
Thus, it is non-trivial to optimize the box parameters with respect to any objective.

\subsection{Differentiable attention map editing}
\label{sec:edit}

To address this, we introduce a differentiable attention map editing method that does not rely on replacement.
Specifically, 
we use the Gaussian edit map $\mM_{\mathcal{G}}$ as a starting point, make its borders smooth so that it is differentiable.
We then propose to edit without relying on 
the binary masks $\mM_{S}$ and $\mM_{T}$.

\paragraph{Smooth masks.}
To prevent the discontinuities shown in \cref{fig:mask_edit}, we smooth out the borders of the Gaussian edit map $\mM_{\mathcal{G}}$, with 1D smooth step functions, both in the horizontal and vertical directions.
Formally, we write
\begin{equation}
\mM_{\mathcal{B}} = \mM_{\mathcal{G}} \odot \mM_{x} \odot \mM_{y}
,
\end{equation}
where $\mM_{x}$ and $\mM_{y}$ are the smooth step functions defined as
\begin{equation}
    \begin{aligned}
        \mM_{x} &= \sig\left(\frac{I_u - b_l}{\kappa}\right) \odot \sig\left(\frac{b_r - I_u}{\kappa}\right) \\
        \mM_{y} &= \sig\left(\frac{I_v - b_t}{\kappa}\right) \odot \sig\left(\frac{b_b - I_v}{\kappa}\right)
    \end{aligned}
    ,
\end{equation}
where $\kappa$ controls the strength of the smooth edge transition, \da{$\sig$ denotes the standard sigmoid function,} and  ${I_u,I_v} 
\in H \times W$. 
Formally, strength $\kappa = \lambda_{edge} \sqrt{h^2 + w^2}$, is calculated as a fraction of the bounding box's diagonal length, where $h, w$ are the height and width of the box. 
In practice, we set $\lambda_{edge}=0.03$. 
We show an example of our attention map edit
in \cref{fig:mask_edit}d.

\paragraph{Differentiable editing.}  
It is important to note that differentiability becomes \da{limited as long as the discrete mask $\mM_{*}$ is utilized.  The first term in Eq. (2) does not provide a useful gradient for optimizing the box coordinates, since the binary mask $\mM_{*}$ represents a hard selection. As a result, the derivative of $\mM_{*}$ with respect to the box coordinates is zero almost everywhere and undefined at the mask boundaries.

Although, the second term may at first glance appear differentiable, it is still evaluated under multiplication by $\mM_{*}$. Since Eq. (2) involves a product between $\mM_{*}$ and $\mM_{\mathcal{G}}$, applying the product rule yields one term involving the derivative of $\mM_{*}$, which vanishes almost everywhere, and another term involving derivatives of $\mM_{\mathcal{G}}$. While the latter term is non-zero, it only affects how values are distributed within the current box and provides limited gradient signal for moving or resizing the box itself.

Though gradients (or subgradients) may flow through the selected values, they do not provide a faithful signal for learning the selection boundaries without an explicit relaxation~\cite{xie2020differentiable}.} 
In contrast, our masks $\mM_{\mathcal{B}}$ are fully differentiable and of similar shape as the binary ones, and can thus be used instead of $\mM_{*}$ in \cref{eq:trailblazer_attention_injection}.
We thus write:\\

\begin{equation}
    \begin{aligned}
        \bar{\mA}_{*} &= \underbrace{\mA_{*} \odot \left(\hparam{w}(1 - \mM_{\mathcal{B}}) + \mM_{\mathcal{B}}\right)}_{\text{weakening outside}} + \underbrace{\hparam{s}\mM_{\mathcal{B}}}_{\text{strengthening inside}}
    \end{aligned}
    ,
    \label{eq:our_injection}
\end{equation}
where again, $\hparam{w}$ and $\hparam{s}$ are the weakening and strengthening factors, respectively, which we set empirically as $\hparam{w}{=}0.001$ and $\hparam{s}{=}0.15$ same as in the case of \trailblazer.
While the difference between \cref{eq:our_injection} and \cref{eq:trailblazer_attention_injection} is subtle,  \cref{eq:our_injection} has no discrete borders, and is thus safe to differentiate with respect to the box parameters.

\subsection{Optimizing bounding boxes to align with attention maps}
\label{sec:opt}

With the attention map editing now being differentiable with respect to the box parameters, we can optimize the bounding box such that the attention map of the next layer is maximized within the box.\footnote{Note that this alone causes the model to highly
ignore outside the bounding box, which we balance; for ease in explanation, we omit this for now.}
The difficulty in doing so, however, is how to define what is a good bounding box edit.
Our hypothesis is that a good edit would be one that is `easy' for the neural network to follow, thus being close to what the neural network itself would do.
This is because both attention map edits in \cref{sec:edit}, are heuristically designed and there is no guarantee that such edit would not derail the video generation process.
In fact, as shown earlier in \cref{fig:teaser}, this does happen.

We thus propose to look into how the edited attention maps propagate through the network layers, and optimize the bounding boxes such that this propagation follows the user's intent.
Specifically, without loss of generality, 
\da{let us denote the transformation induced by the next attention layer as $f$, the `values' in the typical attention operation as $\mV$, } 
we can then write
\begin{equation}
    (\mA_{*}^{l+1}, \mV_{*}^{l+1}) = f(\bar{\mA}_{*}^{l}, \mV_{*}^{l})
    ,
\end{equation}
where the superscript $l$ denotes the layer index, the subscript $*$ again denotes both spatial and temporal, and $\bar{\mA}_{*}^{l}$ is the edited attention map of layer $l$.
We then look at $\mA_{*}^{l+1}$, which now represents how the network is utilizing edited attention, and aim to encourage the attention to be within the user-specified bounding box mask $\mM_{*}$.

Inspired by \da{cross-attention guidance}~\citep{chen2024training}
, we define the loss to encourage that the sum of all attention values $\sum \mA_{*}^{l+1}$ is maximized within the box. 
This can be expressed in various ways, but as the magnitude of attention may differ from one layer to another, we define it such that the sum of the attention values solely come from within the box, that is $\sum \mA_{*}^{l+1} \approx \sum \mA_{*}^{l+1}\odot\mM_{*}^{l+1}$.
\da{The summation is taken over spatial positions and selected tokens; the resulting scalar is squared and averaged over the batch and across layers.} We thus write our loss function as
\begin{equation}
    \loss{attn} = \left\| 1 -  \frac{ 
        \sum \mA_{*}^{l+1}\odot\mM_{*}^{l+1}
    }{
        \sum \mA_{*}^{l+1}
    } \right\|_2^2
    ,
    \label{eq:attn_loss}
\end{equation}
Note that this loss then scales between 0 and 1, with 0 indicating the attention map is purely concentrated to be within the bounding box. 

\paragraph{Preventing attention from completely ignoring outside the box.}
Solely relying on the maximization of attention inside the bounding box causes the model to
ignore the background,
resulting in poor generation quality and less alignment to the prompt. Thus, we introduce a balancing loss, which allows
our method to attend to the outside region, leading to an effective whole generation. 
We write
\begin{equation}
    \loss{$\neg$attn} = \left\| 1-\frac{ 
        \sum \mA_{*}^{l+1}\odot\left(1 - \mM_{*}^{l+1}\right)
    }{
        \sum \mA_{*}^{l+1}
    } \right\|_2^2
    ,
    \label{eq:neg_attn_loss}
\end{equation}
which now enforces outside of the box to retain attention.

\paragraph{Regularizing to remain close to user intent.}
As we wish to preserve as much of the user intent as possible, we
regularize to keep the box close to the original one.
We write
\begin{equation}
    \loss{reg} = \left\| \vb - \vb_{\text{user}} \right\|_2^2
    ,
    \label{eq:reg_loss}
\end{equation}
where $\vb$ would be the optimized bounding box, and $\vb_{\text{user}}$ the original box provided by the user.

\paragraph{Final optimization objective.}
The final loss is then,
\begin{equation}
    \loss{total} = \loss{attn} + \hparam{$\neg$attn}\loss{$\neg$attn} + \hparam{reg} \loss{reg}
    ,
    \label{eq:total_loss}
\end{equation}
where $\hparam{$\neg$attn}$ and $\hparam{reg}$ are the regularization strength, which we empirically set to $\hparam{reg}{=}0.1\times\sqrt{A}$, where $A$ is the number of pixels in the image and $\hparam{$\neg$attn}{=}10$.

\subsection{Implementation details}

We implement our method in PyTorch~\cite{pytorch} based on the official code of \trailblazer.
We further apply our method from \trailblazer to \turbo, a different video generator.
For a fair comparison, we use the same video length of 24 frames, and the default setting for all other hyperparameters associated with \trailblazer.
Because of our resource constraints---memory of our GPU cards is 32GB---we use a resolution of 256 $\times$ 320 and 320 $\times$ 320 for \turbo and \trailblazer, respectively. 
\da{We also use 320 $\times$ 320 resolution for the Peekaboo baseline, and the native resolution 320 $\times$ 512 for Freetraj baseline.}
We find that $\hparam{s}{=}0.3$ works well for the \turbo backbone.
Following \citet{ma2024trailblazer}, we only edit the spatial attention map. 
We run 5 optimization iterations for 5 editing steps,
leading to 25 iterations in total. 
We use the Adam optimizer~\cite{KingBa15} for box adjustments \da{and provide further details on our optimization procedure in the supplemental material.}

\section{Experiments}

\subsection{Experimental setup}

\begin{figure}
\centering
\fbox{
\includegraphics[width=0.95\linewidth]{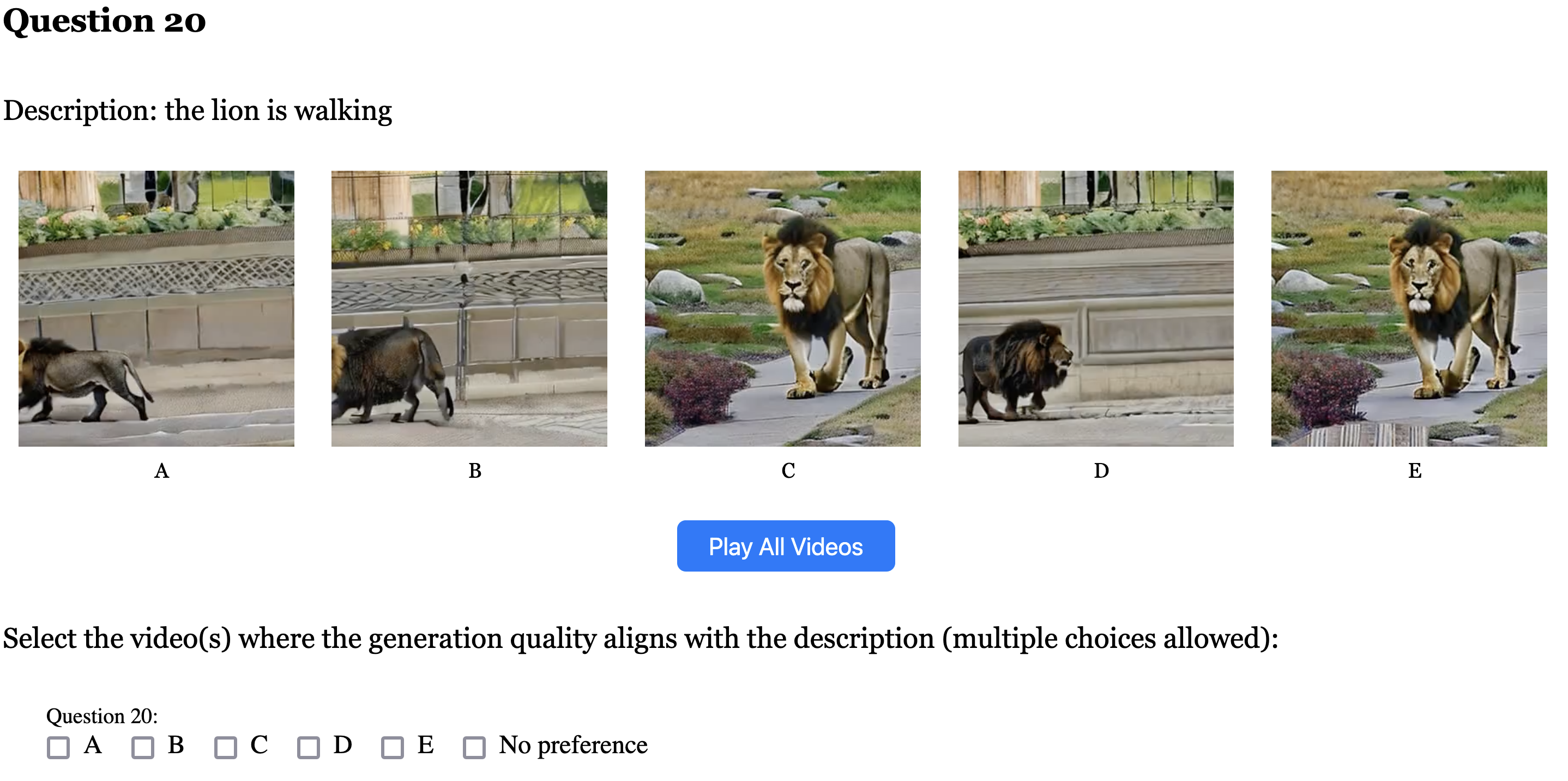}
}
\caption{
    {\bf User study interface -- }
        We show a sample of the user study interface for prompt `the lion is walking'.
        Users are asked to select their preference over a set of five video generations, provided in a random order.
        Users are allowed to select multiple choices or no preference.
        In this example, only the generation results are shown without user control to isolate quality vs. control.
        For this question, C and E look preferable,
        whereas the control is located at the bottom-left where the lion is at A, B, and D.
        C and E completely ignore user control, yet generate a preferred view of the prompt.
        As we are interested in controlled generation, we consider both answers to the quality preference and the trajectory faithfulness when evaluating the quality of generations.
    }
    \label{fig:user_study_sample}
\end{figure}%

Evaluating controllable T2V models is challenging as generated videos lack ground truth and quality metrics only form a proxy to human-perceived quality.
It is also non-trivial to generate large and diverse,
input control trajectories without violating physics, as the generation outcome is highly sensitive to the input control, as we will demonstrate.
For the former, we run a user study alongside human-preference metrics.
For the latter, we use control trajectories from animal videos.
In the following subsections, we will first discuss the experimental setup, including the dataset and how we create the control trajectories. 
We also discuss the baselines, the quantitative metrics, and the user study protocol.

\paragraph{Dataset and bounding-box trajectories.}
As in \citet{ma2024trailblazer}, we use the Animal Kingdom dataset~\cite{ng2022animal} as a reference dataset for actual videos.
This dataset contains several wild animals such as cheetah and hippotamus, with corresponding text describing activity, \eg, \textit{the cheetah is running}. 
This dataset contains 18,744 video clips.

We keep only video clips with a single object using NLTK library~\cite{bird2009nltk}, and moving verbs indicating motion. 
We then use OWL large-scale open vocabulary detector~\cite{minderer2022simple} to detect object boxes, PySceneDetect~\cite{PySceneDetect} library to split scenes to extract trajectories, \ie, sequence of bounding boxes.
To obtain trajectories that are useful, we drop videos that have less than two frames with the object being detected, or with only two detections across scene cuts.
Also, videos that show discontinuous trajectories such as abrupt camera motions or frame cuts are dropped. These are trajectories that have Intersection over Union (IoU) between the consecutive frames being less than 50.
The above results in 1,980 videos (157 unique animals) for training and 526 videos (96 unique animals) for testing. We interpolate detected boxes for the missing frames to obtain a complete trajectory over the video. 
We further filter out trajectories that are of boxes that are too small, \ie, maximum box width and height smaller than 10\% of the width and height of the image, and then randomly sample an interval of 24 frames from the trajectory.
This results in 377 trajectories from the training set that can be used for method design and validation, 
and 226 trajectories exclusively for testing.

\paragraph{Baselines.}
We compare our method against Peekaboo~\cite{jain2024peekaboo}, showcasing the quality improvement of our method.
We also compare against \trailblazer, with its original backbone and we further adapt it to a different
recent text-to-video model \turbo.
We also compare our method against different variations of our method, specifically, our method without box optimization, our method without \cref{eq:neg_attn_loss}, and using our optimized boxes with \trailblazer.

\begin{figure*}[!t]
    \centering
    \newcommand{\vtextheight}{3.2cm}
    \newcommand{\imgw}{0.22\textwidth}
    \setlength{\tabcolsep}{2pt}
    \begin{tabular}{ccccc}
        & Peekaboo (1) & Ours (1) & Peekaboo (2) & Ours (2) \\

        {\rotatebox[origin=l]{90}{\parbox[b]{\vtextheight}{\centering 
        \footnotesize\texttt{The wild dog is running}}}} &
        \includegraphics[width=\imgw]{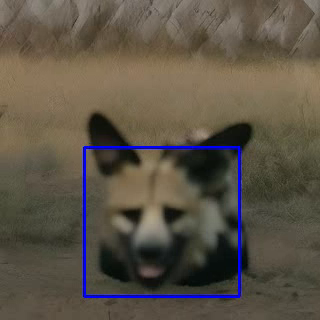} &
        \includegraphics[width=\imgw]{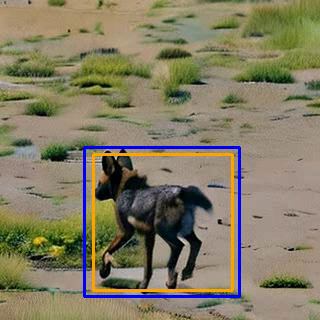} &
        \includegraphics[width=\imgw]{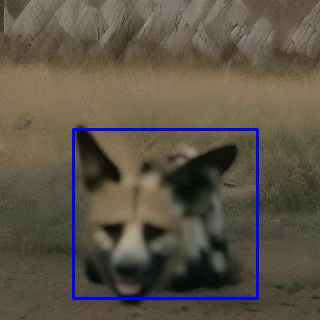} &
        \includegraphics[width=\imgw]{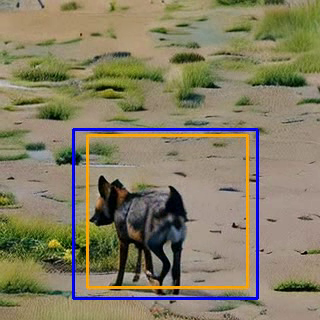} \\

        {\rotatebox[origin=l]{90}{\parbox[b]{\vtextheight}{\centering 
        \footnotesize\texttt{The firebrat insect is moving}}}} &
        \includegraphics[width=\imgw]{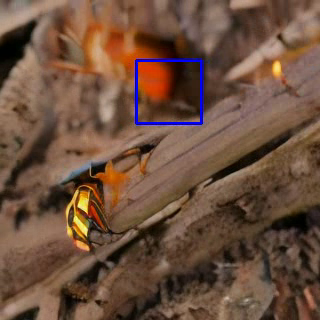} &
        \includegraphics[width=\imgw]{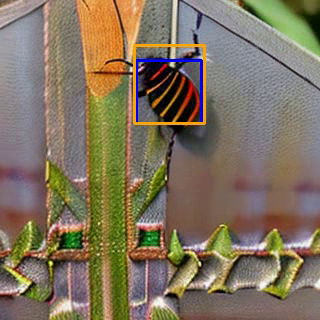} &
        \includegraphics[width=\imgw]{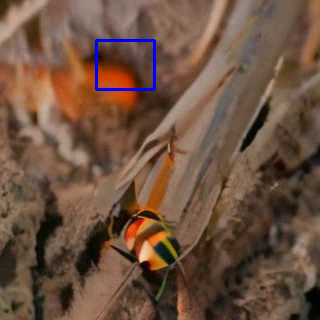} &
        \includegraphics[width=\imgw]{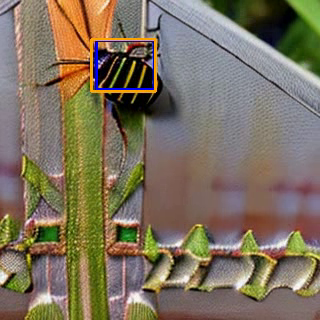} \\
    \end{tabular}
    \vspace{-0.5em}
    \caption{
        {\bf Qualitative results comparing \peekaboo and \da{our method (full)}. -- }
        \small Each row shows two representative frames per method (left to right: Peekaboo, Ours). Our method yields better generation quality and improved alignment with the text prompt.
    }
    \label{fig:ours_vs_peekaboo}
\end{figure*}

\begin{figure*}[!t]
    \centering
    \newcommand{\vtextheight}{3.5cm}
    \newcommand{\vtextheighttv}{3.0cm}
    \newcommand{\imgw}{0.23\textwidth}
    \setlength{\tabcolsep}{2pt}
    \begin{tabular}{cccccc}
        & Baselines & Ours w/o Box Opt. & Baselines + Our boxes & 
        \textbf{Our method} \\
        {\rotatebox[origin=l]{90}{\parbox[b]{\vtextheighttv}{\centering 
        \footnotesize\texttt{The hyena is walking 
        }}}} &
        \includegraphics[width=\imgw]{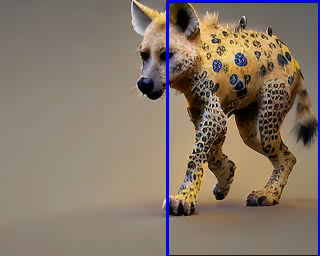} &
        \includegraphics[width=\imgw]{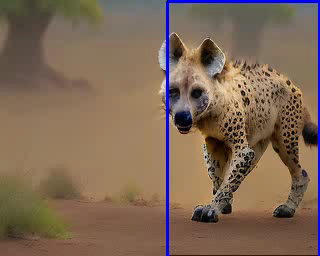} &
        \includegraphics[width=\imgw]{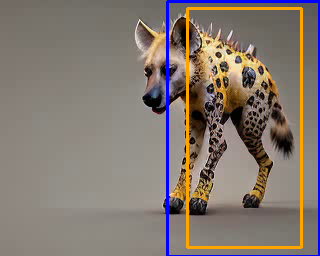} &
        \includegraphics[width=\imgw]{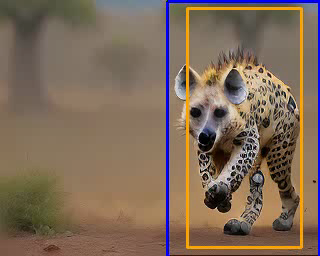} \\

        {\rotatebox[origin=l]{90}{\parbox[b]{\vtextheight}{\centering \footnotesize\texttt{The marine iguana\\ is walking 
        }}}} &
        \includegraphics[width=\imgw]{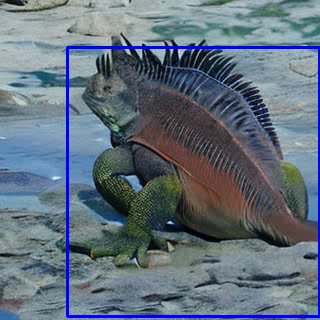} &
        \includegraphics[width=\imgw]{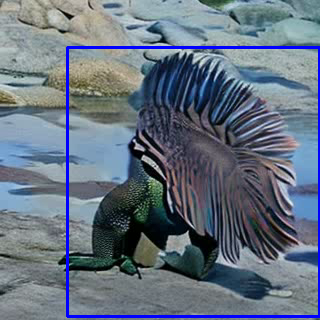} &
        \includegraphics[width=\imgw]{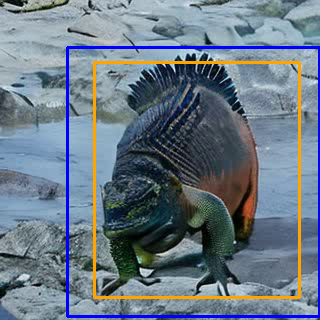} &
        \includegraphics[width=\imgw]{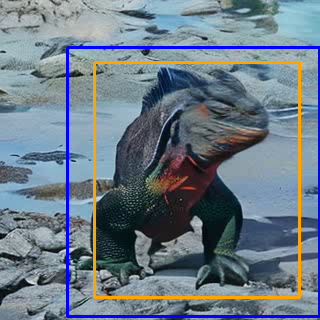} \\

    \end{tabular}
    \vspace{-0.5em}
    \caption{
        {\bf Qualitative results with \turbo (top row) and \trailblazer (bottom row) backbones. -- }
            We display the original user control in \textcolor{blue}{blue} and our optimized boxes in \textcolor{orange}{orange}.
            As shown, ours provides improved rendering quality and adherence to control and the prompt, shown in the top and bottom row.
            See video results for best viewing.
    }
    \label{fig:qualitative}
\end{figure*}

\paragraph{Quantitative
metrics.}
We quantify the performance of each method using human-preference alignment metrics.
We report PickScore~\cite{kirstain2023pick} and HPS v2 (Human Preference Score)~\cite{wu2023human}. We also report the mean intersection over union (mIOU) between the detected objects via an open-vocabulary detector~\cite{minderer2022simple} and the user control, using the animal kingdom dataset as reference.

\paragraph{User study evaluation}.
We also conduct a user study 
with 20 participants selecting their preference over a set of 5 different methods; see \cref{fig:user_study_sample}.
Users were asked to perform two different tasks: (1) to evaluate the quality of the generations without being provided any bounding-box annotations to solely evaluate quality; and (2) to evaluate the adherence to the bounding box.
These were provided as two separate questions and the questions were grouped by tasks.
Multiple choices were allowed, including no preference.
Each participant was asked to answer 40 questions (20 for \turbo and 20 for \trailblazer)
which takes about 20--30 minutes.

To rule out generations that completely failed due to the limitations of the video diffusion model, we chose only those which provide an IoU value of at least 60\% for \emph{at least one of the methods} being compared.
Candidate questions were then selected randomly and anonymized.
A final set with overall quality is used for the study.
This random selection was fixed for all participants.

Finally, as the first task ignores the  user control completely (following user study design), the preference recorded for this tasks can be irrelevant to the controlled generation task as shown in \cref{fig:user_study_sample}.
Thus, we mark as preferred quality only when user selected the video for both tasks.
This study was approved by the Institutional Review Board.

\subsection{Results}

\begin{table}[!t]
    \centering
    \setlength{\tabcolsep}{8pt}
    \resizebox{\linewidth}{!}{%
    \begin{tabular}{@{}l|c c c@{}}
    \toprule
    \textbf{Model} & \textbf{PickScore} $\uparrow$ & \textbf{HPSv2} $\uparrow$ & \textbf{mIOU} $\uparrow$ \\
    \midrule
    \trailblazer & 0.244 & 0.222 & \cellcolor{bestcell} 0.37 \\
    \textbf{Our boxes} + Trailblazer backbone & \cellcolor{bestcell} 0.257 & \cellcolor{secondbestcell} 0.223 & \cellcolor{secondbestcell} 0.36 \\
    \textbf{Our method} w/o Box Opt. & 0.243 & 0.221 & \cellcolor{bestcell} 0.37 \\
    \textbf{Our method} (full) & \cellcolor{bestcell} 0.257 & \cellcolor{bestcell} 0.225 & \cellcolor{bestcell} 0.37 \\
    \midrule
    \midrule
    \peekaboo & 0.125 & 0.189 & 0.30 \\
    \trailblazer & 0.146 & 0.222 & \cellcolor{secondbestcell} 0.37 \\
    \freetraj & 0.178 & 0.223 & 0.34 \\
    Trailblazer + T2V-Turbo backbone & \cellcolor{secondbestcell} 0.234 & \cellcolor{secondbestcell} 0.253 & \cellcolor{bestcell} 0.41 \\
    \textbf{Our method} using T2V-Turbo
     backbone &
    \cellcolor{bestcell}  0.317 & \cellcolor{bestcell} 0.263 & \cellcolor{bestcell} 0.41 \\
    \bottomrule
    \end{tabular}
    }
    \caption{
        {\bf Quantitative results with human-preference and control metrics -- }
        Our full method outperforms baseline and demonstrates consistent preference
        across different architectures, while achieving competitive control.
    }
    \vspace{-1em}
    \label{tab:hps_pickscore_quant}
\end{table}

\begin{figure}
\begin{center}%
\includegraphics[width=0.7\linewidth]{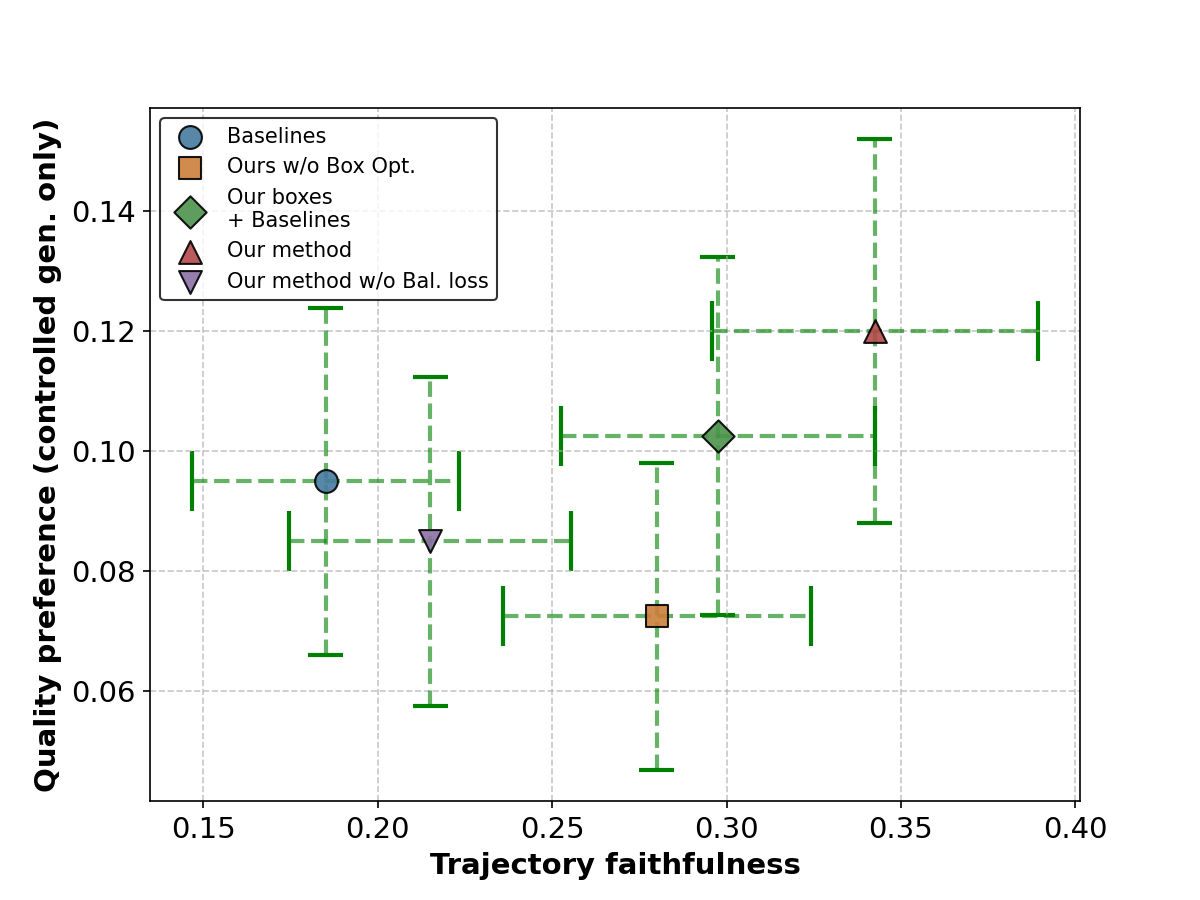}%
\end{center}%
\vspace{-1.5em}
\caption{
    {\bf User study results -- }
    We report the user preference for
    trajectory faithfulness on the x-axis, and quality of generation (for generations that adhere to control) on the y-axis.
    The 95\% confidence intervals are drawn as green dashed lines.
    Our method significantly outperforms \trailblazer with both backbones. 
    }%
    \label{fig:user_study}
\end{figure}%

\paragraph{Quantitative results.}
We report the standard quantitative results in \cref{tab:hps_pickscore_quant}. Overall, our method demonstrates consistent improvements across different architectures \cite{ma2024trailblazer, li2024t2v} (top and bottom row).
Also, our method outperforms baselines such as \peekaboo, \trailblazer, and \da{\freetraj} in terms of 
human preference scores using PickScore and HPSv2 (bottom row). 
Additionally, \textit{"Our boxes + \trailblazer"} validates the benefits of applying our adjusted boxes (top row). 
In terms of control accuracy \da{(mIOU)}, the results show that our method matches the same performance as the baselines. Though the benefit of adjustments is well highlighted in the user studies, this results reveals that our adjustments do not degrade control. 

\begin{figure*}[!t]
    \centering
    \newcommand{\vtextheight}{3.2cm}
    \newcommand{\imgH}{2.5cm}          %
    \setlength{\tabcolsep}{2.2pt}        %

    \newcommand{\imgcell}[1]{%
      \includegraphics[height=\imgH]{#1}%
    }

    \begin{tabular}{ccccc}
        & Freetraj (1) & Ours (1) & Freetraj (2) & Ours (2) \\

        {\rotatebox[origin=l]{90}{\parbox[b]{\vtextheight}{\centering
        \footnotesize\texttt{The drosophila melanogaster fruit fly is moving}}}} &
        \imgcell{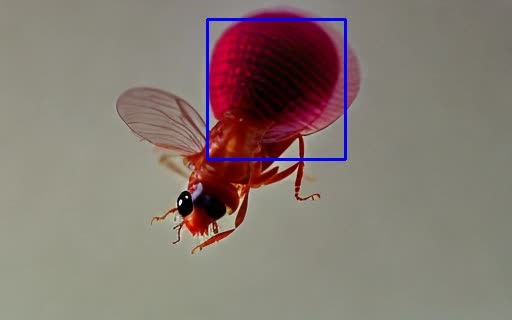} &
        \imgcell{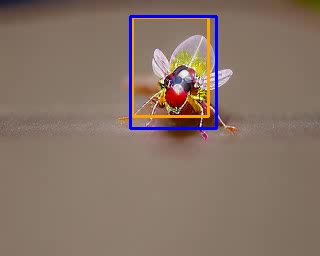} &
        \imgcell{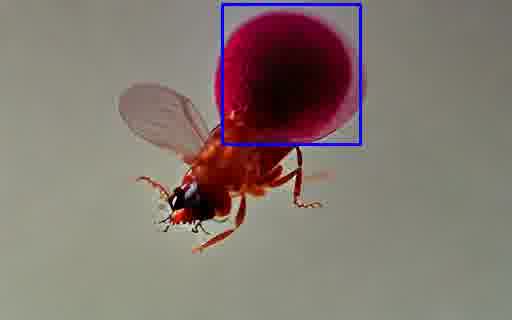} &
        \imgcell{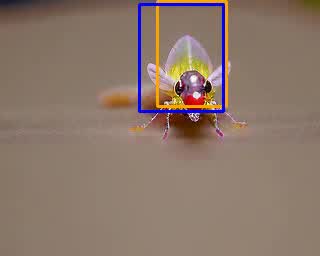} \\

        {\rotatebox[origin=l]{90}{\parbox[b]{\vtextheight}{\centering
        \footnotesize\texttt{The wild dog is running}}}} &
        \imgcell{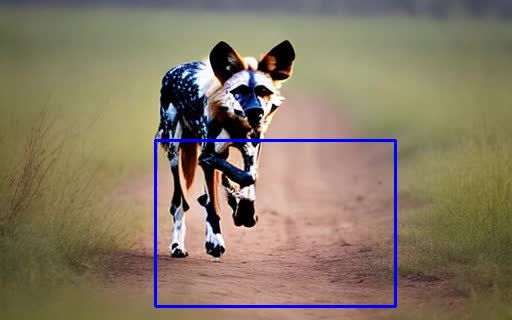} &
        \imgcell{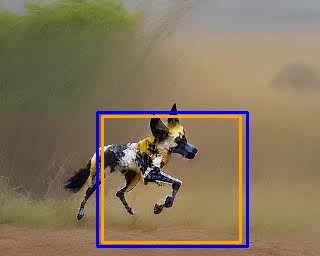} &
        \imgcell{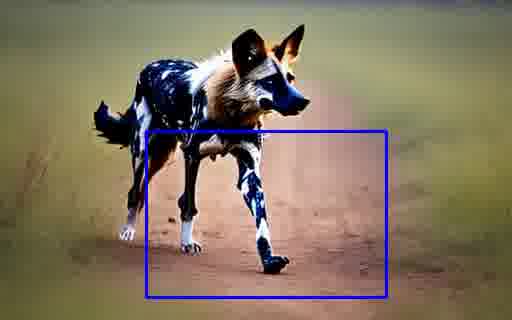} &
        \imgcell{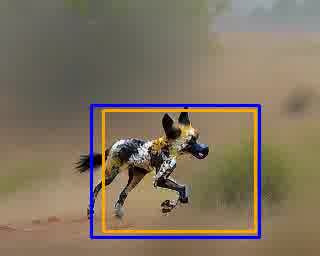} \\
    \end{tabular}

    \vspace{-0.5em}
    \caption{
        {\bf Qualitative results comparing \freetraj and our method on \turbo.}
        \small Frames are shown at a common display height (aspect ratio preserved); frame resolutions differ (Freetraj: $320{\times}512$, Ours: $256{\times}320$). Ours still delivers better generation quality and adherence to control. 
    }
    \label{fig:ours-t2v_vs_freetraj}
\end{figure*}

\begin{figure*}
    \centering
    \newcommand{\vtextheight}{3.5cm}
    \newcommand{\imgw}{0.21\textwidth}
    \setlength{\tabcolsep}{2pt}
    \begin{tabular}{cccccc}
        & Frame 1 & Frame 4 & Frame 16 & 
        Frame 24 \\
        {\rotatebox[origin=l]{90}{\parbox[b]{\vtextheight}{\centering  \trailblazer}}} &
        \includegraphics[width=\imgw]{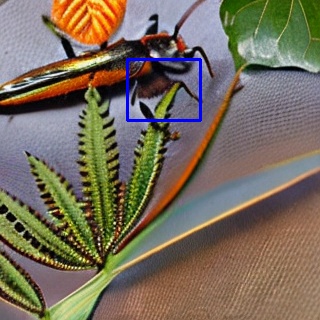} &
        \includegraphics[width=\imgw]{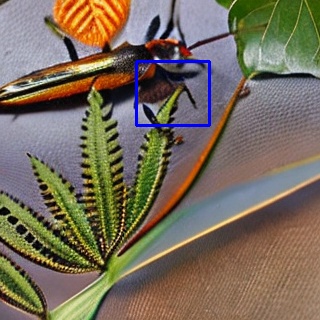} &
        \includegraphics[width=\imgw]{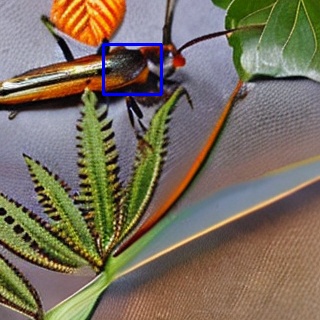} &
        \includegraphics[width=\imgw]{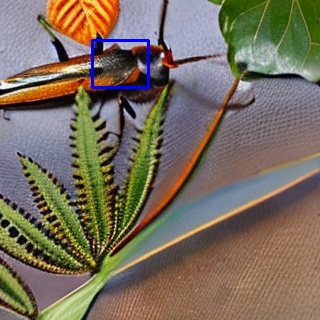} \\
        
        {\rotatebox[origin=l]{90}{\parbox[b]{\vtextheight}{\centering \textbf{Our method}}}} &
        \includegraphics[width=\imgw]{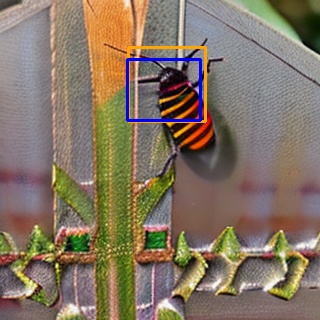} &
        \includegraphics[width=\imgw]{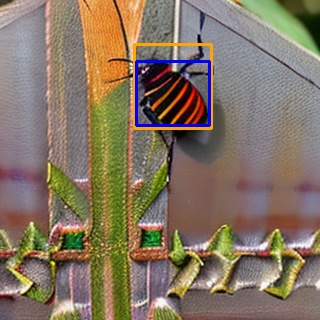} &
        \includegraphics[width=\imgw]{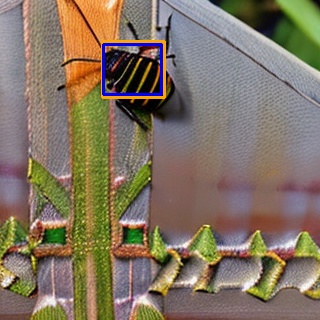} &
        \includegraphics[width=\imgw]{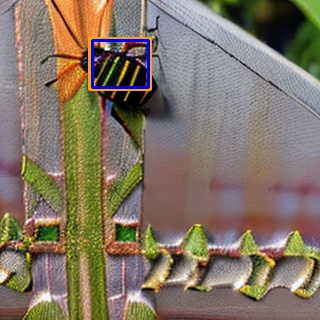} \\

        {\rotatebox[origin=l]{90}{\parbox[b]{\vtextheight}{\centering \trailblazer}}} &
        \includegraphics[width=\imgw]{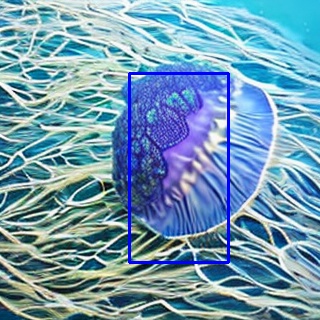} &
        \includegraphics[width=\imgw]{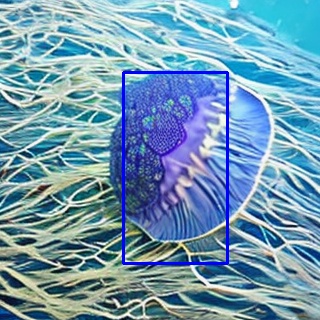} &
        \includegraphics[width=\imgw]{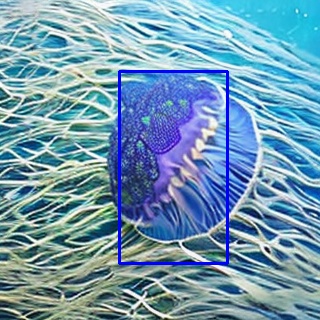} &
        \includegraphics[width=\imgw]{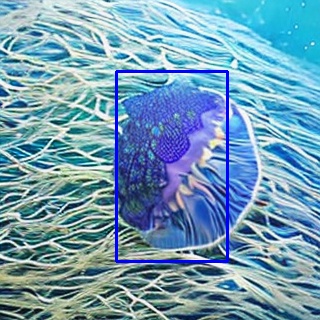} \\
        
        {\rotatebox[origin=l]{90}{\parbox[b]{\vtextheight}{\centering \textbf{Our method}}}} &
        \includegraphics[width=\imgw]{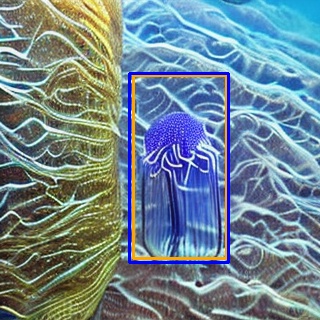} &
        \includegraphics[width=\imgw]{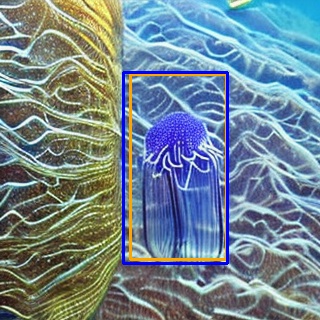} &
        \includegraphics[width=\imgw]{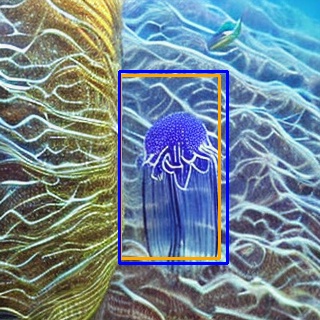} &
        \includegraphics[width=\imgw]{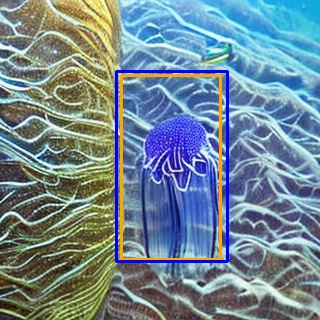} \\

    \end{tabular}
    \caption{
        {\bf Additional qualitative results with \trailblazer -- }
        We show additional examples of our edits.
        As shown, ours is more consistent with user intent and of higher quality.
        The prompt for the top 2 rows is \textit{"The firebrat insect is moving"}, while the prompt for the bottom 2 rows is \textit{"The jellyfish is swimming"}.
        Video results are available in the supplementary local HTML page.
    }
    \label{fig:supp_qualitative_tbl}
\end{figure*}

\begin{figure*}
    \centering
    \newcommand{\vtextheight}{3.5cm}
    \newcommand{\imgw}{0.21\textwidth}
    \setlength{\tabcolsep}{2pt}
    \begin{tabular}{cccccc}
        & Frame 1 & Frame 4 & Frame 16 & 
        Frame 24 \\
        {\rotatebox[origin=l]{90}{\parbox[b]{\vtextheight}{\centering \trailblazer on \turbo }}} &
        \includegraphics[width=\imgw]{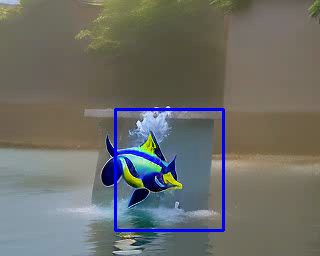} &
        \includegraphics[width=\imgw]{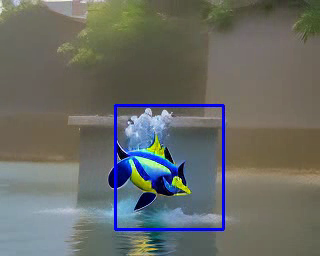} &
        \includegraphics[width=\imgw]{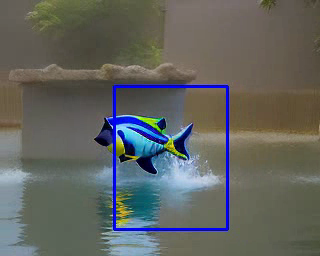} &
        \includegraphics[width=\imgw]{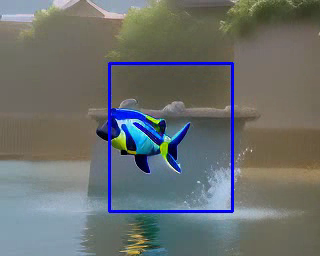} \\
        
        {\rotatebox[origin=l]{90}{\parbox[b]{\vtextheight}{\centering \textbf{Our method on \turbo}}}} &
        \includegraphics[width=\imgw]{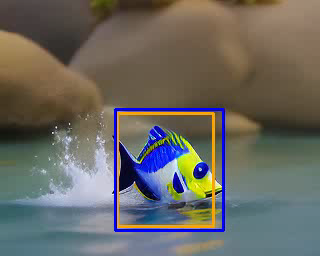} &
        \includegraphics[width=\imgw]{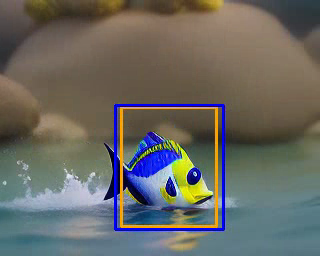} &
        \includegraphics[width=\imgw]{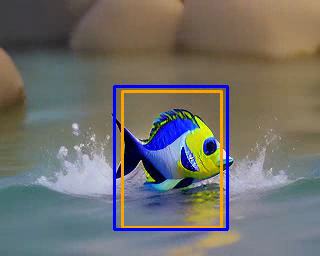} &
        \includegraphics[width=\imgw]{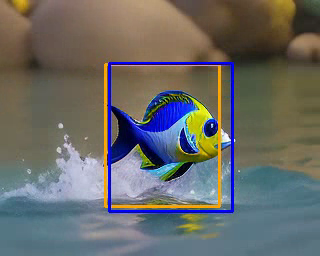} \\

        {\rotatebox[origin=l]{90}{\parbox[b]{\vtextheight}{\centering \trailblazer on \turbo}}} &
        \includegraphics[width=\imgw]{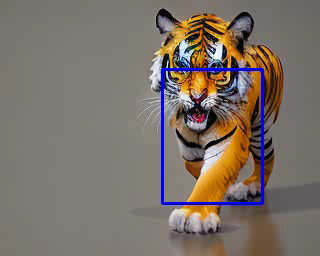} &
        \includegraphics[width=\imgw]{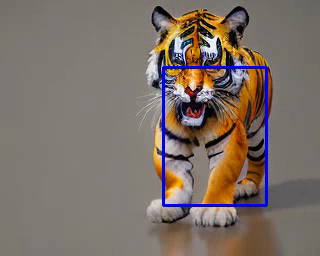} &
        \includegraphics[width=\imgw]{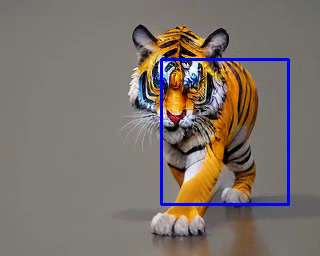} &
        \includegraphics[width=\imgw]{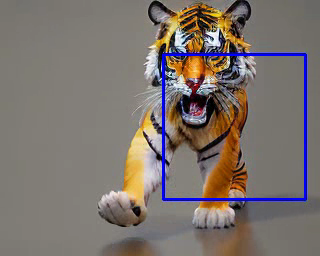} \\
        
        {\rotatebox[origin=l]{90}{\parbox[b]{\vtextheight}{\centering \textbf{Our method on \turbo}}}} &
        \includegraphics[width=\imgw]{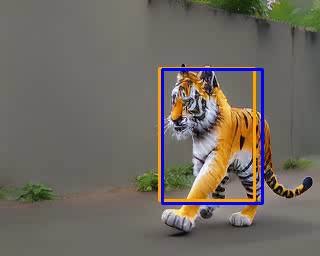} &
        \includegraphics[width=\imgw]{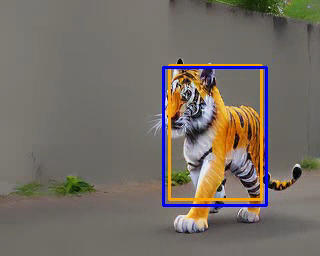} &
        \includegraphics[width=\imgw]{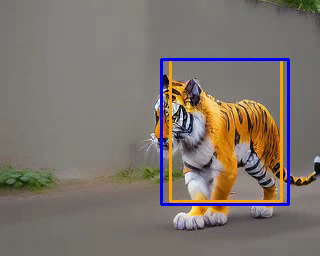} &
        \includegraphics[width=\imgw]{fig/supp_qualitative/t2vturbo/tiger/optim/tiger_motion_29_video_with_bboxes.0000_0015.png} \\

    \end{tabular}
    \caption{
        {\bf Additional qualitative results with \turbo backbone ~-- } We show additional examples of our edits. Ours show more consistency with user intent and is of higher quality within the box. The prompt for the top 2 rows is \textit{"The surgeonfish is swimming"}, while the prompt for the bottom 2 rows is \textit{"The tiger is walking"}.
        Video results are also available in the supplementary local HTML page.
    }
    \label{fig:supp_qualitative_t2vturbo}
\end{figure*}

\paragraph{Qualitative results.}

We show qualitative examples of video frames generated by our method compared to baselines in \cref{fig:ours_vs_peekaboo}, \cref{fig:qualitative}, and \da{\cref{fig:ours-t2v_vs_freetraj}}. Notably, our method \da{performs} significantly better than \peekaboo \da{and \freetraj}, \da{in terms of generation quality and adherence to control, in \cref{fig:ours_vs_peekaboo} and \da{\cref{fig:ours-t2v_vs_freetraj}} respectively.}

\da{Also, when} compared with T2V-Turbo and Trailblazer backbones, our method, courtesy of adjustments, produces better generation that adhere with the prompt in \cref{fig:qualitative}. This is especially visible in the second~row, where generation follows motion in the prompt \textit{"The marine iguana is walking"}.
Interestingly, applying our adjusted
bounding boxes also provides better prompt following for the baseline method.
Though this benefit is not always transferable, as the baseline still uses a discrete editing strategy. 
Best performance with adjustment is achieved with our method. Finally, in \cref{fig:supp_qualitative_tbl} and \cref{fig:supp_qualitative_t2vturbo}, we show additional qualitative results of our method compared to baselines.

\paragraph{User evaluation.}
We report user study results in \cref{fig:user_study}. 
As shown, our method is the most preferred by a large margin.
It is also worth noting that using our adjusted
boxes also benefits \trailblazer, as shown by the slight increase in preference.
We hypothesize that this is because our boxes are optimized for the same underlying model.
This further demonstrates that small adjustments matter, not only for the method
that we have introduced, but also for the base model.
Finally, Ours w/o balancing loss 
(\ie,~\cref{eq:neg_attn_loss} 
being left out) show that taking the overall generation into account is effective. 
Simply optimizing the bounding box, without considering the whole generation
results in a middle ground between the baseline and the non-optimized version. 
\da{However}, when both are enabled, we can achieve a significant performance increase.

\section{Conclusion}

In this work, we demonstrated that small adjustments to user-provided bounding boxes can lead to significant improvements in controlled video generation. 
By optimizing the bounding boxes to better align with the internal attention maps of video diffusion models while maintaining proximity to user inputs, we achieved both higher quality generations and better adherence to control signals.
Through extensive experiments and user studies, we validated that our simple yet effective approach outperforms existing methods for controlled video generation. 
We believe our findings open up new possibilities for improving user control in text-to-video generation by considering how control signals can be optimized to work better with pretrained models.

\begin{figure}
    \centering
    \begin{subfigure}[b]{0.32\linewidth}
        \includegraphics[width=\linewidth]{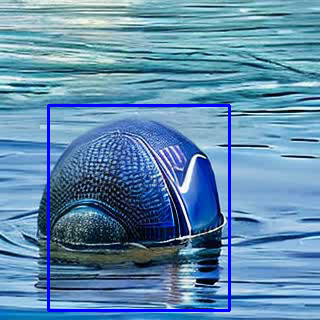}
        \caption{\trailblazer}\label{fig:failure_a}
    \end{subfigure}
    \hspace{0.05\linewidth} %
    \begin{subfigure}[b]{0.32\linewidth}
        \includegraphics[width=\linewidth]{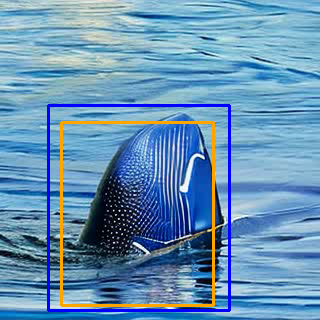}
        \caption{Our method}\label{fig:failure_b}
    \end{subfigure}
    \caption{
        {\bf Failure cases -- }
            We show example failure cases when the video model fails to generate the desired output for the prompt \texttt{the orca is swimming}.
            Here, regardless of the bounding box input, the model is unable to generate the content as it has no knowledge of the target prompt.
            We hypothesize that more advanced video models, which are being made increasingly available~\cite{genmo, wan2025wan}, will alleviate this problem. 
    }
    \label{fig:failure_case}
\end{figure}

\paragraph{Limitations and future work} 
A limitation of our method is that the quality of generated images can be bound by the underlying video model, as shown in \cref{fig:failure_case}.
With the rapid progress in this area, we believe this limitation will be alleviated naturally.
While in theory our method can be applied to other forms of control, we have only validated our idea to bounding boxes. \da{But beyond bounding boxes, our work highlights a broader principle: control signals themselves are often misaligned with a model’s internal representations, and modest optimization can substantially improve outcomes. This perspective naturally extends to future work on optimizing other conditional inputs, such as trajectories, sketches, or depth cues.}
Finally, our method requires 
partial back-propagation during optimization, thus slows down generation.
With the \turbo backbone, our generations take one minute and 37 seconds, \da{while Trailblazer + original T2V Turbo takes 48 seconds, both on an NVIDIA RTX A6000.}
While our edits bring significant enhancements, optimizing this would also be an interesting extension. 
\da{Conclusively}, our main novelty demonstrates that small adjustments matter, and we therefore wish to draw the attention of the community to the sensitivity of control signals and hope that our work will serve as a foundational ground.

\da{\paragraph{Broader impact statement}
Our method is applicable beyond animals, and generations can be abused with malicious intent. For example, controlling the movement of celebrities for personal gain. We discourage such practices and hope that providing proper awareness can alleviate or minimize variations of such intent.}

\da{\paragraph{Acknowledgement}
This work was supported by Borealis AI through the Borealis AI Global Fellowship Award. 
We acknowledge the computational resources provided by the University of British Columbia and thank the volunteers who participated in the user study.}

\bibliography{macros_eric, main}
\bibliographystyle{tmlr}

\clearpage
\setcounter{page}{1}
\maketitlesupplementary
\appendix

We provide additional information to the main content, including an ablation study, failure cases, further implementation details, \da{and results with complex patterns and difficult trajectories. We also include more results, in terms of videos, as well as an interactive slider that visualizes how internal attention maps evolve during optimization, on the project webpage.}

\section{Ablation study}
\paragraph{Regularization strength for deviation.}
We ablate the impact of the regularizer strength in \cref{eq:total_loss} for how much the box deviates from user intent. 
We show our results with \trailblazer in \cref{fig:reg_strength}. 
As shown, stronger regularization leads to bounding boxes that are more faithful to the user input, but their generation results are poor in quality. 
Too low values create too much deviation.
$\hparam{reg}{=}0.1$ strikes a good balance between user intent and generation quality.

\paragraph{Linear trajectories.}
We also show our method applied to linear trajectories to simulate a user input.
In \cref{fig:linear3d_motion_t2v}, we use \turbo with our method and provide a bounding-box that linearly interpolates between a start and an end bounding box.
As shown, 
our method provides enhancement.

\begin{table}[!ht]
    \centering
    \setlength{\tabcolsep}{4pt}
    \resizebox{\linewidth}{!}{%
    \begin{tabular}{@{}l cccc@{}}
    \toprule
    Attention map & Attention reshaped  & \# Edit channels  & Edited copy (lower channels) & Layer info 
    \\
    \midrule
     240, 1600, 77 & 240, 40, 40, 77 & 120 & 120, 40, 40, 77 & down\_blocks.0.0.0 \\

    240, 1600, 77 & 240, 40, 40, 77 & 120 & 120, 40, 40, 77 & down\_blocks.0.1.0 \\
    
    480, 400, 77 & 480, 20, 20, 77 & 240 & 240, 20, 20, 77 & down\_blocks.1.0.0 \\
    
    480, 400, 77 & 480, 20, 20, 77 & 240 & 240, 20, 20, 77 & down\_blocks.1.1.0 \\
    
    960, 100, 77 & 960, 10, 10, 77 & 480 & 480, 10, 10, 77 & down\_blocks.2.0.0 \\
    
    960, 100, 77 & 960, 10, 10, 77 & 480 & 480, 10, 10, 77 & down\_blocks.2.1.0 \\
    
    960, 25, 77 & 960, 5, 5, 77 & 480 & 480, 5, 5, 77 & mid\_block.0.0 \\
    
    960, 100, 77 & 960, 10, 10, 77 & 480 & 480, 10, 10, 77 & up\_blocks.1.0.0 \\
    
    960, 100, 77 & 960, 10, 10, 77 & 480 & 480, 10, 10, 77 & up\_blocks.1.1.0 \\
    
    960, 100, 77 & 960, 10, 10, 77 & 480 & 480, 10, 10, 77 & up\_blocks.1.2.0 \\

     480, 400, 77 & 480, 20, 20, 77 & 240 & 240, 20, 20, 77 & up\_blocks.2.0.0 \\
    
    480, 400, 77 & 480, 20, 20, 77 & 240 & 240, 20, 20, 77 & up\_blocks.2.1.0 \\
    
    480, 400, 77 & 480, 20, 20, 77 & 240 & 240, 20, 20, 77 & up\_blocks.2.2.0 \\
    
    240, 1600, 77 & 240, 40, 40, 77 & 120 & 120, 40, 40, 77 & up\_blocks.3.0.0 \\
    
    240, 1600, 77 & 240, 40, 40, 77 & 120 & 120, 40, 40, 77 & up\_blocks.3.1.0 \\

    240, 1600, 77 & 240, 40, 40, 77 & 120 & 120, 40, 40, 77 & up\_blocks.3.2.0 \\

    \bottomrule
    \end{tabular}
    }
    \caption{
        {\bf Edited cross attention maps -- }
        As in \trailblazer, we edit only the lower channels across all layers.
        Each layer follows the spatial cross-attention naming where \textit{down\_blocks.0.0.0} can be fully written as \textit{\textbf{down\_blocks}.\textbf{0}.attentions.\textbf{0}.transformer\_blocks.\textbf{0}.attn2}.
    }
    \label{tab:cross_attn_layers}
\end{table}

\section{Failure cases}

Additionally, 11\% and 33\% of the users did not have a preference for the quality and trajectory faithfulness questions respectively. 
This can be attributed to the limitations of the underlying video model both in terms of inability to faithfully generate according to the prompt and at desired locations.
Still, with more advanced video models becoming increasingly available~\cite{genmo, wan2025wan}, we expect this problem to be mitigated in the future.

\section{Further implementation details}
\label{sec:further_max}

\begin{figure*}[!ht]
    \centering
    \newcommand{\vtextheight}{3.5cm}
    \newcommand{\imgw}{0.23\textwidth}
    \setlength{\tabcolsep}{2pt}
    \begin{tabular}{cccccc}
        {\rotatebox[origin=l]{90}{\parbox[b]{\vtextheight}{\centering \footnotesize\texttt{The wolf is exploring}}}} &
        \includegraphics[width=\imgw]{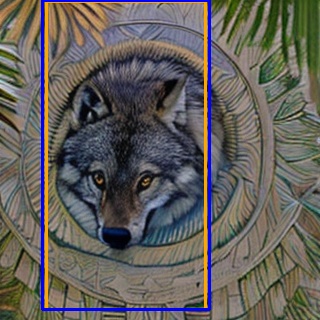} &
        \includegraphics[width=\imgw]{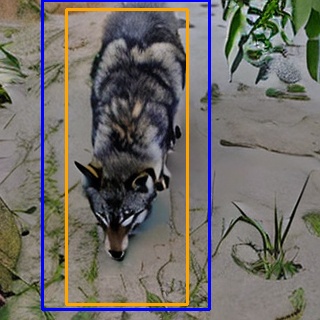} &
        \includegraphics[width=\imgw]{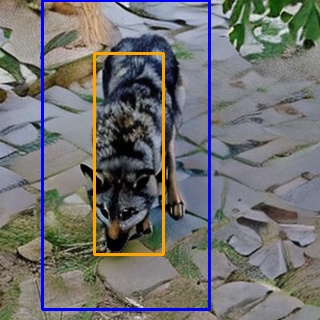} &
        \includegraphics[width=\imgw]{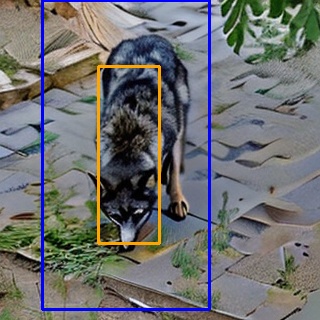} \\
        & 1.0 & 0.1 & 0.01 & 0.001 
    \end{tabular}
    \caption{
        {\bf Effect of regularization strength/penalty -- } We show an example frame with original  bounding boxes as \textcolor{blue}{blue} and the adjusted boxes as \textcolor{orange}{orange}.
        A higher penalty (1.0) 
        allows less movement, 
        while too loose penalty (0.001) causes generation to stray away from user intent.
        We set the penalty to 0.1 for all our experiments, which strikes a good balance.
    }
    \label{fig:reg_strength}
\end{figure*}

\paragraph{Optimization procedure}
We present our optimization procedure in Algorithm~\ref{optim_algo1} and show how optimization interleaves with the diffusion process.
\begin{algorithm}
\caption{Trajectory optimization}
\begin{algorithmic}[1]
\State \textbf{Input:} Video diffusion model $\Theta_b$, decoder $D$, initial latents $\vz$, prompt embedding $\vp$,
denoise steps $K$, spatial edit steps $K_S$, temporal edit steps $K_T$, inner optimization steps $O$, learning rate $\mathcal{u}$, and trajectory boxes $\mathcal{B}$

\For{$i \leftarrow 1$ to $K$}
    \State $r \leftarrow 0$ \Comment{inner optimization step counter}
    \State $t \leftarrow \mathrm{TimestepSchedule}(i)$ 

    \While{$(i < K_S$ or $i < K_T)$ and $r < O$}
        \State /* extract attention maps and noise prediction */
        \State $(\mA_S,\mA_T,\epsilon_t) \leftarrow \Theta_b(\vp,\vz,t)$

        \State /* Apply editing (Eq. 6), compute loss (Eq. 11) and gradients */
        \State $(\tilde{\mA}_S,\tilde{\mA}_T) \leftarrow \mathrm{edit}(\mA_S,\mA_T)$ \Comment{Eq. 6}
        \State $\mathcal{L} \leftarrow \mathcal{L}(\tilde{\mA}_S,\tilde{\mA}_T,\mathcal{B})$ \Comment{aggregated over frames and layers; Eq. 11}
        \State $g \leftarrow \nabla_{\mathcal{B}} \mathcal{L}$

        \State /* Update bounding boxes */
        \State $\mathcal{B} \leftarrow \mathrm{AdamStep}(\mathcal{B}, g; \mathcal{u})$

        \State /* Project boxes to valid 
        boundaries */
        \State $\mathcal{B} \leftarrow \mathrm{Proj}(\mathcal{B})$

        \State $r \leftarrow r + 1$
    \EndWhile

    \State /* standard denoise step (no editing) */
    \State $\epsilon_t \leftarrow \Theta_b(\vp,\vz,t)$
    \State $\vz \leftarrow \mathrm{SchedulerStep}(\vz,\epsilon_t,t)$
\EndFor

\State \textbf{return} $D(\vz)$
\end{algorithmic}
\label{optim_algo1}
\end{algorithm}

\paragraph{Keeping bounding boxes within images.}
To prevent optimization from making bounding boxes go out of images, we simply clip the optimized bounding box coordinates.

\paragraph{Additional hyperparameters.}
\label{sec:hyper_param}
Besides the hyperparameters noted in the main paper, for others, we follow \trailblazer to be comparable.
Specifically for \trailblazer, we use Zeroscope v2 576w as the underlying text-to-video model,\footnote{Zeroscope v2 576w model. \url{https://huggingface.co/cerspense/zeroscope_v2_576w}} and the same DPM Solver\footnote{Cheng Lu \etal, DPM-Solver: A Fast ODE Solver for Diffusion Probabilistic Model Sampling in Around 10 Steps, \textit{Advances in Neural Information Processing Systems}, 2022} for both methods, with a guidance scale of $9.0$.
We run $40$ denoising steps, applying the editing (and our optimization) on the first five steps. 
For Peekaboo baseline, we run the same number of denoising steps with the default number of frozen steps set to 2. 
\da{For Freetraj baseline, we run with $50$ denoising steps.} We also use their DDIM eta of $0$, with $6$ edit steps, and set the video length to 24 frames, matching the input trajectory.

For \turbo, we use the official implementation, with the default parameters and $16$ denoising steps. We apply the edits, both of our method and \trailblazer for only the first five steps.

\paragraph{Control metric.}
For computing the mIoU, we simply choose, among the detected bounding boxes, one that matches most closely to the user control bounding box. 
Note that this is different from \trailblazer which relies on the object detection score.  Our strategy takes into account that generated videos are not constrained to generate only a single object, as 
there may be distractors. We found that \trailblazer strategy of using the detection scores does get easily distracted by distractors, thus providing an unfair comparison. Our strategy fixes this.

\paragraph{Model architecture.}
\label{sec:model_param}
For \trailblazer, we use the U-Net architecture of Zeroscope.
An initial gaussian noise $z_{t}$ with shape 1 $\times$ 4 $\times$ F $\times$ 40 $\times$ 40 is generated where F $=$ 24 frames.
We edit only the spatial attention with a target video resolution of 320 $\times$ 320. 
The corresponding attention maps are computed from the queries and keys, while the total number of tokens is 77~\cite{radford2021learning}; each token has a dimension of 1024. 
The progression of the edited attention maps with their respective dimensions is detailed in~\cref{tab:cross_attn_layers}.

For \turbo, the base model utilizes the same U-Net-style architecture as \trailblazer, hence we edit in the same way as for \trailblazer.
We use PyTorch~\cite{pytorch} gradient checkpointing in our implementations.

\def\subfigwidth{0.19}

\begin{figure*}[!t]
    \centering
    \setlength{\tabcolsep}{1pt}
    \begin{tabular}{cccc}
        \multicolumn{4}{c}{\footnotesize\texttt{The cuttlefish is swimming}} \\
        \begin{subfigure}[b]{\subfigwidth\linewidth}
            \includegraphics[width=\linewidth]{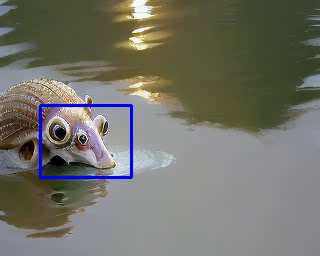}
            \caption{\centering \trailblazer\ on \\ \turbo}
        \end{subfigure} &
        \begin{subfigure}[b]{\subfigwidth\linewidth}
            \includegraphics[width=\linewidth]{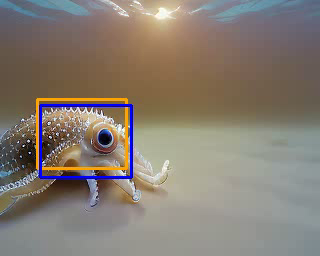}
            \caption{\centering Our method \\ on \\ \turbo}
        \end{subfigure} &
        \begin{subfigure}[b]{\subfigwidth\linewidth}
            \includegraphics[width=\linewidth]{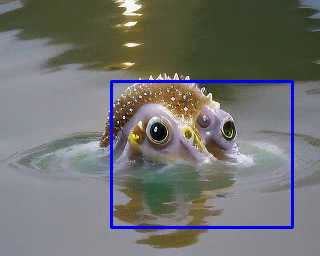}
            \caption{\centering \trailblazer\ on \\ \turbo}
        \end{subfigure} &
        \begin{subfigure}[b]{\subfigwidth\linewidth}
            \includegraphics[width=\linewidth]{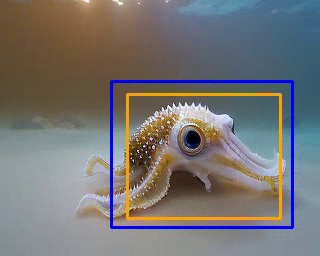}
            \caption{\centering Our method \\ on \\ \turbo}
        \end{subfigure} \\[-2pt] 
        \multicolumn{4}{c}{\footnotesize\texttt{The horse is running}} \\
        \begin{subfigure}[b]{\subfigwidth\linewidth}
            \includegraphics[width=\linewidth]{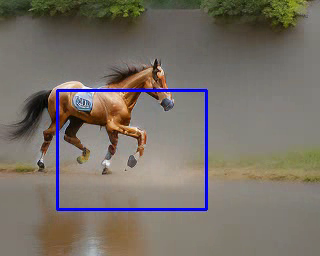}
            \caption{\centering \trailblazer\ on \\ \turbo}
        \end{subfigure} &
        \begin{subfigure}[b]{\subfigwidth\linewidth}
            \includegraphics[width=\linewidth]{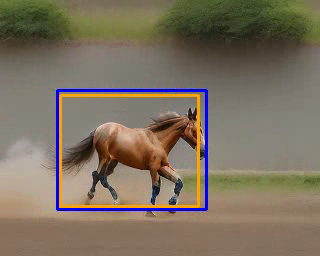}
            \caption{\centering Our method \\ on \\ \turbo}
        \end{subfigure} &
        \begin{subfigure}[b]{\subfigwidth\linewidth}
            \includegraphics[width=\linewidth]{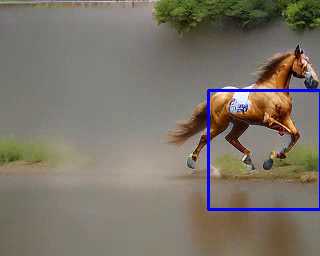}
            \caption{\centering \trailblazer\ on \\ \turbo}
        \end{subfigure} &
        \begin{subfigure}[b]{\subfigwidth\linewidth}
            \includegraphics[width=\linewidth]{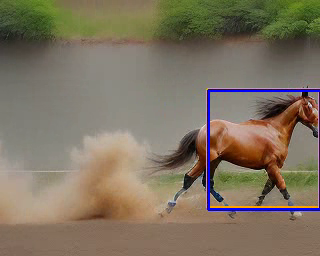}
            \caption{\centering Our method \\ on \\ \turbo}
        \end{subfigure}
    \end{tabular}
    \caption{{\bf Application to perspective and linear trajectories (top and bottom respectively) -- } We show example frames of applying our method to \turbo with a trajectory that linearly interpolates two bounding boxes (start and end).
    Ours still delivers improvements, both in terms of generation quality within the bounding box and adherence.
    }
    \label{fig:linear3d_motion_t2v}
\end{figure*}

\section{Ablations on component design choices}
\label{sec:further_comp_ablation}

We include comprehensive ablations on key design choices in \cref{tab:deviation_penalty,tab:kernel_normalize,tab:outside_bbox_loss_scale,tab:edge_strength}.

\paragraph{Deviation penalty.} \cref{tab:deviation_penalty} shows that excessive deviation leads to loss of user intent (lower mIoU), while overly restrictive penalties limit quality improvements. A penalty of 0.1 balances control and generation quality.

\paragraph{Smoothing kernel normalization.} \cref{tab:kernel_normalize} demonstrates that normalizing the smoothing kernel to produce consistent peak attention across layers significantly improves PickScore, HPSv2, and mIoU.

\paragraph{Background preservation term.} \cref{tab:outside_bbox_loss_scale} confirms that the background attention term is necessary for maintaining scene coherence.

\paragraph{Edge strength.} \cref{tab:edge_strength} shows that very low edge strength behaves similarly to discontinuous masks (e.g., Trailblazer), degrading quality, while overly strong edges reduce control accuracy. Intermediate values (0.001–0.03) provide the best trade-off.

\begin{table}[!t]
\centering
\small
\begin{tabular}{l l l l}
\hline
$ \alpha$ & \textbf{PickScore} $\uparrow$ & \textbf{HPSv2} $\uparrow$ & \textbf{mIOU} $\uparrow$ \\
\hline
0 & \cellcolor{secondbestcell} 0.210 & \cellcolor{bestcell} 0.230 & 0.31 \\
0.001 & \cellcolor{bestcell} 0.211 & \cellcolor{bestcell} 0.230 & 0.32 \\
0.01 & 0.199 & \cellcolor{secondbestcell} 0.228 & 0.36 \\
0.1 & 0.194 & 0.225 & \cellcolor{bestcell} 0.38 \\
1 & 0.187 & 0.221 & \cellcolor{secondbestcell} 0.37 \\
\hline
\end{tabular}
\caption{Ablation results for deviation penalty term $\hparam{reg}{=}\alpha \times\sqrt{A}$}
\label{tab:deviation_penalty}
\end{table}

\begin{table}[!t]
\centering
\small
\begin{tabular}{l l l l}
\hline
Normalization & \textbf{PickScore} $\uparrow$ & \textbf{HPSv2} $\uparrow$ & \textbf{mIOU} $\uparrow$ \\
\hline
True & \cellcolor{bestcell} 0.509 & \cellcolor{bestcell} 0.225 & \cellcolor{bestcell} 0.38 \\
False & \cellcolor{secondbestcell} 0.491 & \cellcolor{secondbestcell} 0.223 & \cellcolor{bestcell} 0.38 \\
\hline
\end{tabular}
\caption{Ablation results for smoothing kernel normalization}
\label{tab:kernel_normalize}
\end{table}

\begin{table}[!t]
\centering
\small
\begin{tabular}{l l l l}
\hline
$\hparam{$\neg$attn}$ & \textbf{PickScore} $\uparrow$ & \textbf{HPSv2} $\uparrow$ & \textbf{mIOU} $\uparrow$ \\
\hline
0 & 0.228 & 0.219 & 0.36 \\
1 & 0.231 & 0.221 & \cellcolor{bestcell} 0.39 \\
10 & \cellcolor{secondbestcell} 0.256 & \cellcolor{secondbestcell} 0.225 & \cellcolor{secondbestcell} 0.38 \\
100 & \cellcolor{bestcell} 0.285 & \cellcolor{bestcell} 0.228 & 0.31 \\
\hline
\end{tabular}
\caption{Ablation results for background preservation term $\hparam{$\neg$attn}$}
\label{tab:outside_bbox_loss_scale}
\end{table}

\begin{table}[!t]
\centering
\small
\begin{tabular}{l l l l}
\hline
 $\lambda_{edge}$ & \textbf{PickScore} $\uparrow$ & \textbf{HPSv2} $\uparrow$ & \textbf{mIOU} $\uparrow$ \\
\hline
0.0001 & 0.240 & 0.224 & \cellcolor{bestcell} 0.40 \\
0.001 & \cellcolor{secondbestcell} 0.248 & \cellcolor{secondbestcell} 0.226 & \cellcolor{secondbestcell} 0.38 \\
0.03 & 0.243 & 0.225 & \cellcolor{secondbestcell} 0.38 \\
0.1 & \cellcolor{bestcell} 0.268 & \cellcolor{bestcell} 0.232 & 0.36 \\
\hline
\end{tabular}
\caption{Ablation results for edge strength $\lambda_{edge}$}
\label{tab:edge_strength}
\end{table}

\section{Complex patterns and difficult trajectories}
\label{sec:complex_patterns}

\begin{figure*}
    \centering
    \newcommand{\vtextheight}{3.5cm}
    \newcommand{\imgw}{0.23\textwidth}
    \setlength{\tabcolsep}{2pt}
    \begin{tabular}{ccccc}
        & Frame 5 & Frame 9 & Frame 13 \\
        {\rotatebox[origin=l]{90}{\parbox[b]{\vtextheight}{\centering  \trailblazer}}} &
        \includegraphics[width=\imgw]{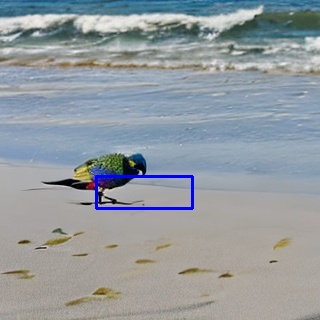} &
        \includegraphics[width=\imgw]{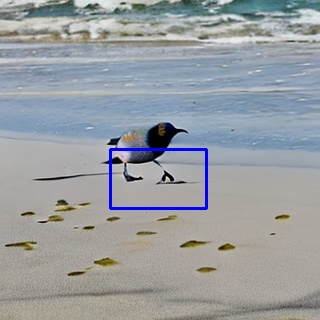} &
        \includegraphics[width=\imgw]{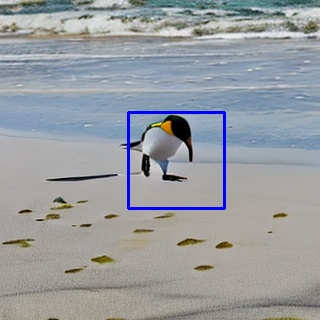} \\
        
        {\rotatebox[origin=l]{90}{\parbox[b]{\vtextheight}{\centering \textbf{Our boxes} + Trailblazer backbone }}} &
        \includegraphics[width=\imgw]{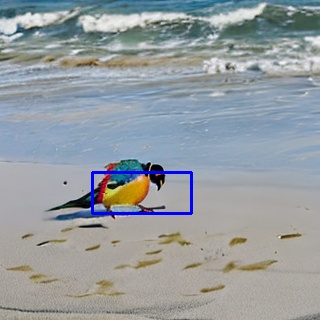} &
        \includegraphics[width=\imgw]{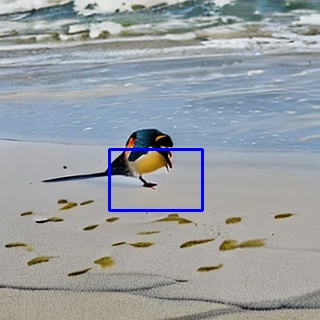} &
        \includegraphics[width=\imgw]{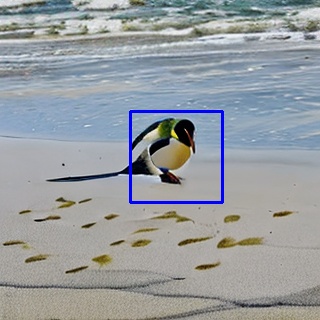} \\
           
    \end{tabular}
    \caption{
        {\bf Additional qualitative results for morphing task (Parrot $\rightarrow$ Penguin)-- } We show transition frames, where our optimized boxes provide benefit for better control of the morphing task.
    }
    \label{fig:morphing_task}
\end{figure*}

\begin{figure*}
    \centering
    \newcommand{\vtextheight}{3.5cm}
    \newcommand{\imgw}{0.18\textwidth}
    \setlength{\tabcolsep}{2pt}
    \begin{tabular}{cccccc}
        & Frame 1 & Frame 4 & Frame 16 
        & Frame 24 
        \\
        {\rotatebox[origin=l]{90}{\parbox[b]{\vtextheight}{\centering  \textbf{Our method}}}} &
        \includegraphics[width=\imgw]{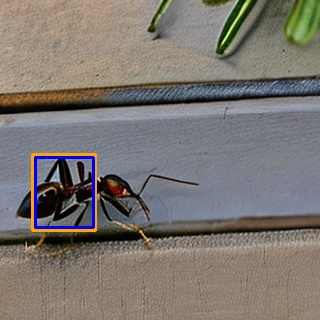} &
        \includegraphics[width=\imgw]{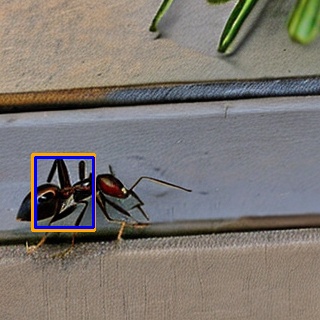} &
        \includegraphics[width=\imgw]{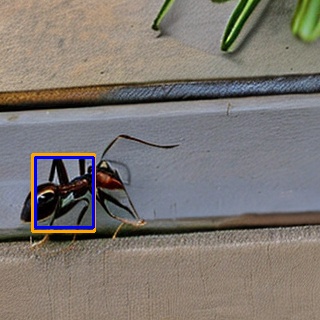} &
        \includegraphics[width=\imgw]{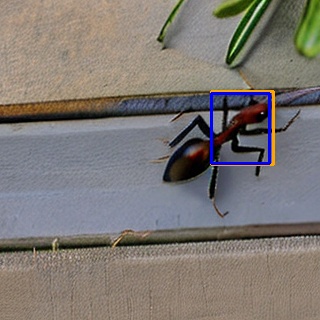} \\
        
        {\rotatebox[origin=l]{90}{\parbox[b]{\vtextheight}{\centering \textbf{Our method}}}} &
        \includegraphics[width=\imgw]{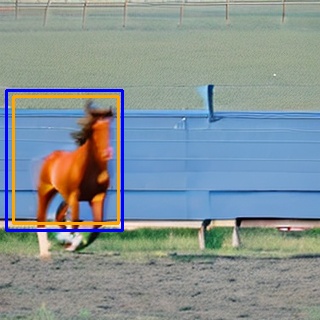} &
        \includegraphics[width=\imgw]{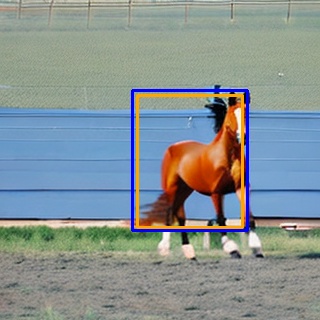} &
        \includegraphics[width=\imgw]{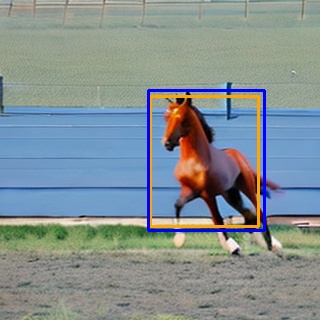} &
        \includegraphics[width=\imgw]{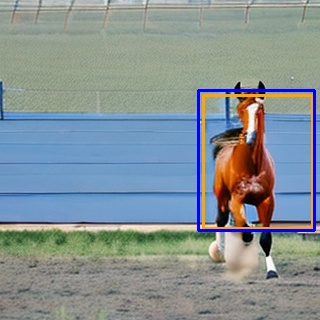} \\

    \end{tabular}
    \caption{
        {\bf Additional qualitative results with stationary-to-move and U-turn trajectories -- }
        We show example frames where our method is applied to complex patterns and preserves difficult motions. Best viewed in motion \da{on the project webpage}.
    }
    \label{fig:stationary_to_move_task}
\end{figure*}

\begin{figure*}
    \centering
    \newcommand{\vtextheight}{3.5cm}
    \newcommand{\imgw}{0.18\textwidth}
    \setlength{\tabcolsep}{2pt}
    \begin{tabular}{cccccc}
        & Frame 1 & Frame 4 & Frame 16 & 
        Frame 24 \\
        
        {\rotatebox[origin=l]{90}{\parbox[b]{\vtextheight}{\centering \textbf{Our method}}}} &
        \includegraphics[width=\imgw]{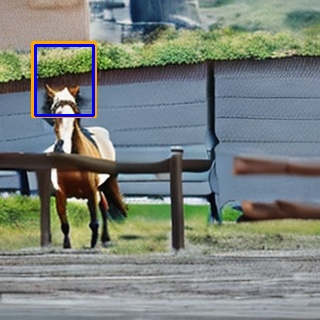} &
        \includegraphics[width=\imgw]{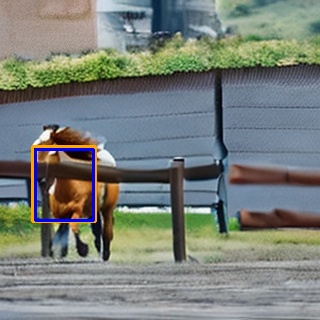} &
        \includegraphics[width=\imgw]{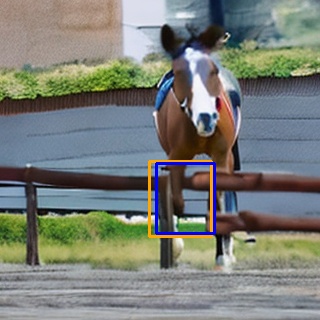} &
        \includegraphics[width=\imgw]{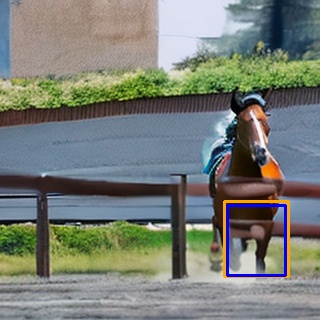} \\

    \end{tabular}
    \caption{
        {\bf Extreme scenario case with very small boxes and in a zig-zag trajectory -- }
        We present an extreme scenario with zig-zag motion using very small boxes, different from the object size. Our method compensates and generates motion that follows user intent. Best viewed in motion \da{on the project webpage}.
    }
    \label{fig:zig_zag_task}
\end{figure*}

We show results for complex patterns and difficult trajectories such as morphing, zig-zag, U-turn, and stationary-to-move in \cref{fig:morphing_task,fig:zig_zag_task,fig:stationary_to_move_task}.

\end{document}